\documentclass[lettersize,journal]{IEEEtran}
\usepackage{amsmath,amsfonts}
\usepackage{algorithmic}
\usepackage{algorithm}
\usepackage{array}
\usepackage[caption=false,font=normalsize,labelfont=sf,textfont=sf]{subfig}
\usepackage{textcomp}
\usepackage{stfloats}
\usepackage{url}
\usepackage{verbatim}
\usepackage{graphicx}
\usepackage{cite}
\usepackage{xcolor}
\usepackage{multirow}
\usepackage{booktabs}
\usepackage[flushleft]{threeparttable} 
\newcommand{\myparagraph}[1]{\vspace{2mm}\noindent\textbf{#1}}

\usepackage{hyperref}
\usepackage{caption}

\usepackage{cuted} 
\begin{document}

\title{Vision Tomography: Towards Portable 3D Bidirectional Face-Skull Inference from Visual Data}

\author{
Zidu Wang, \and 
Meng Xu, \and
Miao Xu, \and 
Hengyuan Ma, \and 
Jiankuo Zhao, \and
Xutao Li, \and  \\
Zhen Lei~\IEEEmembership{Fellow,~IEEE}, \and
Xiangyu Zhu~\IEEEmembership{Senior Member,~IEEE}

\thanks{Zidu Wang, Miao Xu, Jiankuo Zhao, Xutao Li, Zhen Lei, and Xiangyu Zhu are with the State Key Laboratory of Multimodal Artificial Intelligence Systems, Institute of Automation, Chinese Academy of Sciences, and the School of Artificial Intelligence, University of Chinese Academy of Sciences, Beijing, China. Email: wangzidu0705@gmail.com, \{xumiao2021, zhaojiankuo2024, lixutao2025, zhen.lei, xiangyu.zhu\}@ia.ac.cn}
\thanks{Meng Xu and Hengyuan Ma are with Plastic Surgery Hospital, Chinese Academy of Medical Sciences and Peking Union Medical College, Beijing, China. Email: xum8701@gmail.com, mahy@psh.pumc.edu.cn}
\thanks{Miao Xu and Zhen Lei are also with the Centre for Artificial Intelligence and Robotics, Hong Kong Institute of Science \& Innovation, Chinese Academy of Sciences, Hong Kong, China}
\thanks{Correspondence to Dr. Xiangyu Zhu.}

}


\maketitle

\begin{abstract}
Traditional computed tomography (CT) can capture paired 3D face and skull geometry, but it remains costly, fixed in location, and exposes patients to radiation. For applications such as remote diagnostics, digital healthcare, forensic identification, missing-person investigation, and archaeological analysis, there is an urgent need for a portable, safe, and accessible solution to capture 3D skull geometry. In this work, we introduce Vision Tomography (VT), a concept and framework capable of reconstructing paired 3D face-skull geometry from various visual modalities, and also enabling joint manipulation of 3D facial and skull geometry. To achieve this, we first construct a dataset of over 200 samples, where each case includes a CT-based skull, a CT-based face, and a high-fidelity textured face scan. Secondly, we propose a novel dense ray matching registration method that ensures topological consistency across face, skull, and their tissue correspondences. Building on this registration, we introduce the 3D Bidirectional Face-Skull Morphable Model (BFSM) as the foundation of Vision Tomography, which enables shape inference between the face and skull through a shared coefficient space, while also modeling tissue thickness variation to support one-to-many facial reconstructions from the same skull, reflecting individual facial changes such as age or body fat. Finally, we demonstrate the effectiveness of Vision Tomography across various forensic and medical applications, including paired 3D face–skull reconstruction from images or sketches and predicting surgical outcomes with joint face–skull editing. Extensive experiments confirm the robustness and accuracy of our method. Vision Tomography is available at \href{https://github.com/wang-zidu/BFSM}{https://github.com/wang-zidu/BFSM}.
\end{abstract}


\section{Introduction}

\begin{figure*}[t]
\centering
\includegraphics[width=1.0\textwidth]{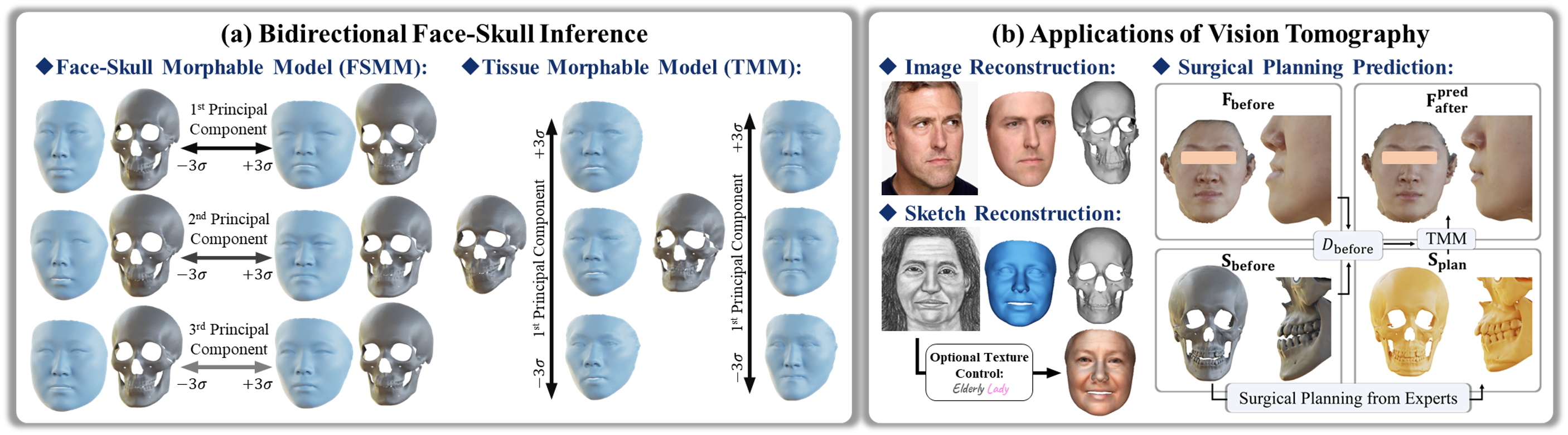}
\captionof{figure}{(a) We introduce the 3D Bidirectional Face-Skull Morphable Model (BFSM) as the foundation of Vision Tomography (VT), which enables shape inference between the face and skull through a shared coefficient space, while also modeling tissue thickness variation to support one-to-many facial reconstructions from the same skull. (b) Applications of Vision Tomography, including 3D face-skull reconstruction from visual data and joint editing of face and skull geometry for surgical planning.
}
\label{teaser}
\end{figure*}

\begin{figure}[t]
\centering
\includegraphics[width=1.0\columnwidth]{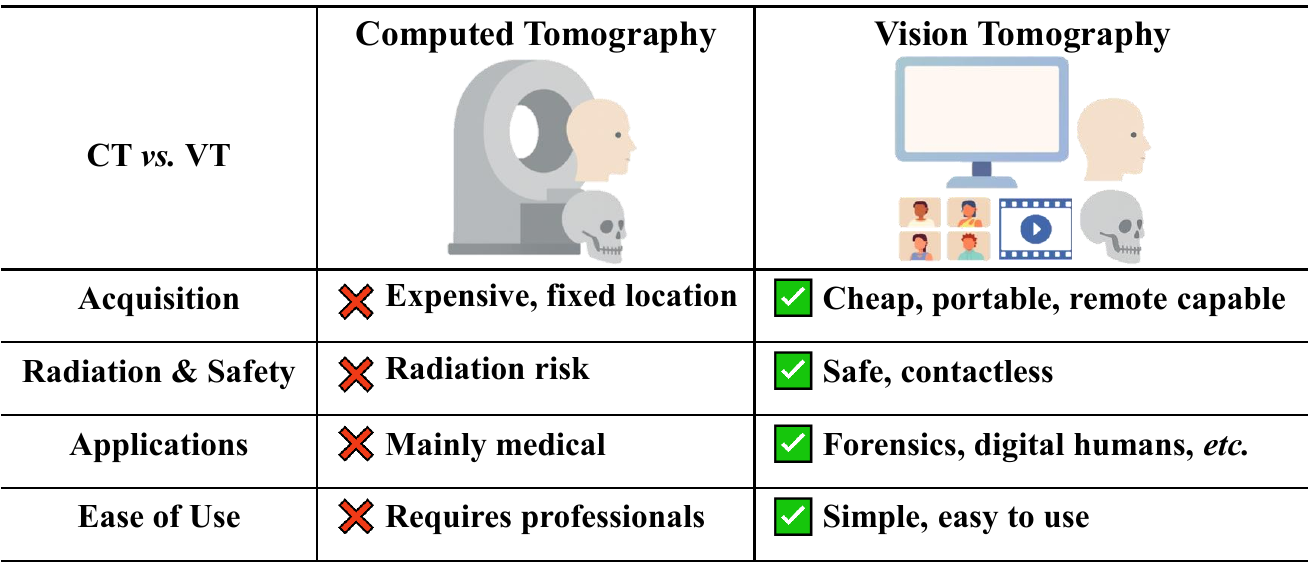}
\caption{A simple comparison between Computed Tomography (CT) and Vision Tomography (VT).}
\label{CT_vs_VT}
\end{figure}

3D bidirectional face–skull inference refers to inferring skull geometry from a given 3D face and vice versa. When coupled with advanced image-based 3D face reconstruction techniques \cite{wang20243d,wang2024s2td,10127617,10506678,DECA:Siggraph2021,deng2019accurate}, it enables anatomically consistent recovery of paired face–skull geometry from visual observations. For forensic identification and archaeological research, non-invasive and portable reconstruction of facial appearance and cranial structure is essential. For telemedicine and aesthetic surgical planning, there is a strong demand for methods that can rapidly infer approximate skull geometry from facial images and predict postoperative facial outcomes from edited skull geometry, providing preliminary anatomical guidance and intuitive visualizations to support clinical decision-making and doctor–patient communication. Such technologies can also help bridge healthcare gaps in underdeveloped regions with limited access to specialized expertise and imaging tools. Despite its broad applicability, paired 3D face–skull geometry is still mainly obtained via computed tomography (CT), which is expensive, location-bound, and involves ionizing radiation. These limitations highlight the urgent demand for a safe, cheap, and accessible solution. To address this, we propose Vision Tomography (VT), a portable framework for reconstructing 3D face and skull from images or sketches, while supporting joint editing of facial and skull geometry, as shown in Figure~\ref{teaser}(b). Figure~\ref{CT_vs_VT} provides a simple comparison between CT and VT. 3D face and skull reconstruction from visual inputs remains challenging, as the internal structure needs to be inferred rather than directly measured. This inference heavily relies on both high-quality paired 3D data and accurate prior geometry models.

Limited paired face-skull data is a major challenge in recent efforts to jointly model face and skull geometry. Some studies rely on datasets with fewer than 100 identities \cite{luthi2009hierarchical, paysan2009face, madsen2018probabilistic,10791299SkulltoFace, achenbach2018multilinear, SCULPTOR}, while common 3D Morphable Models (3DMMs) \cite{blanz1999morphable, blanz2003face, FLAME:SiggraphAsia2017} focus primarily on facial geometry. A few methods \cite{madsen2018probabilistic, Ichim_Phace} construct joint morphable models from unpaired data, which can introduce biases and reduce anatomical accuracy. SCULPTOR \cite{SCULPTOR} is limited to mandibular changes using before-and-after orthognathic surgery data, neglecting maxillary modeling. To overcome the data limitations, we construct a new dataset comprising over 200 high-quality samples. Every sample includes a CT-based skull, a CT-based face, and a high-fidelity textured face scan, making it the most comprehensive dataset of its kind to date, as shown in Figure~\ref{preprocess}(d).

Modeling difficulties arise from both the complex surface geometry of the skull and the challenge of accurately capturing the correspondence between facial soft tissue and underlying bone. Skull deformation for topology registration is more challenging than for the face, as existing non-rigid methods \cite{nicp_github, amberg2007optimal} depend on precise control points to avoid local optima, yet identifying corresponding landmarks between source and target skull data is much harder than on the face. In many cases, skull geometry is non-watertight due to internal holes, which hinders the use of implicit representations such as signed distance fields (SDFs) \cite{mildenhall2020nerf,wang2021neus}. Although several works \cite{yang2024learning, yang2023implicit, yang2022implicit, zoss2019accurate, chandran2024anatomically} explore implicit modeling, they tend to oversimplify skull geometry, which makes them unsuitable for clinical scenarios that require high anatomical fidelity. Furthermore, the modeling of soft tissue between the face and skull remains an underexplored problem. Only a few studies \cite{achenbach2018multilinear, 10791299SkulltoFace, keller2024hit, keller2023skin, keller2022osso, SCULPTOR} have attempted to capture the tissue relationship between face and skull, but limitations such as small paired datasets, sparse correspondences, focus on the mandible or general skeletal modeling hinder accurate joint manipulation of 3D facial and skull geometry. To overcome the modeling difficulties, we propose a novel dense ray matching registration method tailored for face and skull geometry. It fully incorporates the geometrical priors of anatomically aligned face and skull structures, ensuring topological consistency across the face, skull, and their connecting tissues. As shown in Figure~\ref{teaser}(a), leveraging both the anatomical diversity of our dataset and the innovations in registration modeling, we introduce the 3D Bidirectional Face-Skull Morphable Model (BFSM). BFSM comprises two key components: the Face-Skull Morphable Model (FSMM), which enables bidirectional shape inference between the face and skull through a shared latent space, and the Tissue Morphable Model (TMM), which models tissue thickness variation to support one-to-many face reconstructions from a given skull, accounting for individual differences such as aging change and body fat distribution.

We explore applications of Vision Tomography (VT) by leveraging BFSM to reconstruct 3D faces and skulls from visual data and enable joint editing of their geometry. In forensic scenarios, both the 3D face and the corresponding skull can be directly reconstructed from sketches or images using BFSM, enabling identification support and historical investigation. In medical contexts, VT allows preliminary surgical planning without relying on CT scans: patients can capture facial images with a smartphone, reconstruct their craniofacial geometry in 3D, and collaboratively perform synchronous face-skull editing with clinicians online, then visualize the projected post-surgical outcomes on the original face to illustrate surgical changes. These applications of VT provide practical value for digital healthcare, forensic identification, and missing-person investigations, enhancing user experience and professional efficiency through interactive visualization and manipulation. Our main contributions are as follows:

(1) \textbf{Data:} A dataset comprising over 200 samples, with paired CT-based skulls, CT-based faces, and high-fidelity textured face scans, making it the most comprehensive dataset to date.

(2) \textbf{Modeling:} A novel dense ray matching topology registration method that uses anatomically aligned priors to ensure topological consistency across different face-skull pairs and their tissues, leading to a novel 3D Bidirectional Face-Skull Morphable Model (BFSM) that enables bidirectional shape inference between face and skull through a shared latent space and captures tissue thickness variability to support diverse face reconstructions.

(3) \textbf{Application:} Vision Tomography has been demonstrated in applications such as paired face-skull reconstruction from various visual data and joint editing of face and skull geometry for surgical planning, highlighting its potential for diverse applications in healthcare, forensics, \textit{etc}.

\section{Related Work}

\subsection{3D Face and Skull Data}

Large-scale 3D face datasets have become widely available, with recent collections such as FaceVerse \cite{wang2022faceverse}, FaceScape \cite{yang2020facescape}, NPHM \cite{giebenhain2023nphm}, and LYHM \cite{dai2020statisticalLYHM}, each comprising hundreds of identities. In contrast, only a few studies provide 3D skull data \cite{madsen2018probabilistic, achenbach2018multilinear, luthi2009hierarchical, paysan2009face}, and the number of skull identities is substantially smaller. Paired face-skull datasets remain particularly scarce, for example, the LUCY dataset \cite{SCULPTOR} only includes 72 identities with pre- and post-orthognathic operations. Overall, existing paired datasets are limited in scale, lack demographic diversity, restricting their utility for modeling anatomically varied populations.

\subsection{Geometry Deformation and Modeling} 
Deformation-based topology registration refers to non-rigidly deforming a source geometry to match a target geometry. It is fundamental to modeling 3D face–skull geometry, especially for building parametric deformation models. Existing non-rigid deformation methods \cite{nicp_github, amberg2007optimal, porto2021alpaca,sorkine2007rigid}, mostly based on nearest-neighbor strategies, rely heavily on accurate control points to avoid local optima. In practice, only a small number of sparse landmarks are available, and even slight errors can cause significant registration failures.

Combined with topology registration and non-rigid deformation methods, parametric or morphable models are commonly used for 3D human-related modeling. Widely used facial morphable models include the Basel Face Model (BFM) \cite{blanz1999morphable, blanz2003face} and FLAME \cite{FLAME:SiggraphAsia2017}. Several works explore implicit representations \cite{yang2024learning, yang2023implicit, yang2022implicit, zoss2019accurate, chandran2024anatomically} for physically based facial simulation, but often simplify skull geometry to a degree that limits clinical relevance. Other methods \cite{keller2024hit, keller2023skin, keller2022osso, achenbach2018multilinear, 10791299SkulltoFace} attempt to model face-skull relationships jointly, yet they typically overlook specific clinical application needs, which may reinforce biases and limit the inclusivity of healthcare solutions. SCULPTOR \cite{SCULPTOR} relies solely on pre- and post-orthognathic surgery data to model skull-face deformations, biasing the model toward mandibular changes while limiting its ability to capture maxillary deformations. Some works have also explored dental reconstruction \cite{xu2024teethdreamer, li2020learning} or attempted to build implicit parametric morphable dental models \cite{zhang2022implicit}.

\subsection{Face and Skull Reconstruction and Editing}
For image-based reconstruction, many recent methods \cite{wang2025srmhair, wang20243d, DECA:Siggraph2021, SMIRK, deng2019accurate} can capture high-quality 3D facial geometry and expressions, and some can even reconstruct detailed 3D faces from sketches of different styles \cite{wang2024s2td}. However, reconstructing both face and skull geometry from images remains largely unexplored, constraining applications in forensic investigation, archaeological analysis, and digital healthcare. SCULPTOR \cite{SCULPTOR} attempted this, but focused only on mandible reconstruction rather than the entire skull, highlighting the need for anatomically complete face–skull reconstruction methods.

For 3D face and skull geometry reconstruction or mutual inference, several methods have explored neural-network-based geometry transformations. FAST-Net \cite{zhao2023fast} mainly addresses face-to-skull prediction, while skull-to-face reconstruction is less explored, and the evaluation is conducted on a limited dataset. Ma et al. \cite{ma2023bidirectional} uses neural networks to model subject-specific bidirectional transformations between facial and craniofacial skeletal shapes, though it was trained on fewer than 50 samples. Bao et al. \cite{bao2024deep} introduced point-based neural networks (P2P-ConvGC) for bidirectional transformations, while TCFNet \cite{zhang2025tcfnet} proposes a Transformer-based \cite{vaswani2017attention} architecture for similar tasks. These approaches \cite{ma2023bidirectional,bao2024deep,zhang2025tcfnet} typically employ separate but structurally similar networks for face-to-skull and skull-to-face inference, and are mainly limited to point cloud transformations without modeling facial texture. Moreover, implicit neural approaches rely heavily on dataset priors, making them sensitive to dataset bias, and they generally do not account for the one-to-many nature of skull-to-face inference caused by individual variations such as fat distribution, as discussed in prior work \cite{SCULPTOR,10791299SkulltoFace}. 

For 3D face and skull geometry editing, existing digital human editing techniques \cite{potamias2024shapefusion, tarasiou2024locally, foti20223d} predominantly manipulate facial surfaces and do not account for the anatomical relationship between face and skull. Several studies have explored postoperative outcome prediction in craniofacial surgery. GPOSC-Net \cite{kim2025predicting} predicts post-operative cephalometric outcomes from pre-operative lateral cephalograms, but cephalometric representations lack the geometric expressiveness of explicit 3D meshes for visualization and communication. Du et al. \cite{du2024postoperative} focuses on postoperative facial morphology prediction in mandibular defect reconstruction, without explicitly modeling the joint deformation between facial soft tissue and craniofacial skeletal structures, despite their strong correlation in craniofacial surgery. More recent approaches investigate joint face–skull modeling and editing. Yang et al. \cite{yang2025blending} propose a coarse-to-fine framework that combines parametric face modeling (SCULPTOR \cite{SCULPTOR}) with neural network refinement to predict postoperative facial soft tissue changes in orthognathic surgery, while ACMT-Net \cite{fang2024correspondence} maps skeletal deformation to facial geometry via attentive correspondence.

\section{Data}
This section describes our data acquisition process and basic preprocessing steps. Paired 3D face and skull data form the basis for enabling Vision Tomography.

\subsection{Ethics.} 
Ethical approval was obtained from the Institutional Review Board. All participants provided written informed consent. The study adhered to the principles of the \textit{Declaration of Helsinki}.

\subsection{Data Acquisition.} 

We collaborate closely with several experienced clinicians, with over a decade of clinical practice. These experts meticulously review the 3D data for each case to ensure anatomical accuracy and clinical relevance. Under their guidance, we filter out anatomically irrelevant or redundant structures—such as internal cranial tissues or scattered noise artifacts. As shown in Figure~\ref{preprocess} (a) and (b), our dataset is composed of two sources: computed tomography (CT) and 3D textured scanning. The CT data, encompassing both skull $\mathbf{S}_{i}^\mathrm{CT}$ and face $\mathbf{F}_{i}^{\mathrm{CT}}$, are acquired using a \textit{Philips Brilliance iCT 128-slice spiral CT scanner}, with scanning focused on the craniofacial regions. High-resolution textured face scannings $\mathbf{F}_{i}^\mathrm{3D}$ are captured using the \textit{3dMD Face System}. The facial scans cover the area from ear to ear horizontally and from the hairline down to the neck vertically, with texture at a resolution of 5 megapixels. The meshes $\mathbf{S}_{i}^{\mathrm{CT}}$, $\mathbf{F}_{i}^{\mathrm{CT}}$, and $\mathbf{F}_{i}^{\mathrm{3D}}$ represent the CT-based skull, CT-based face, and 3D-scanned face of subject $i$, respectively.

\subsection{Clinician-guided and Automated Alignment.}

\begin{figure}[t]
\centering
\includegraphics[width=1.0\columnwidth]{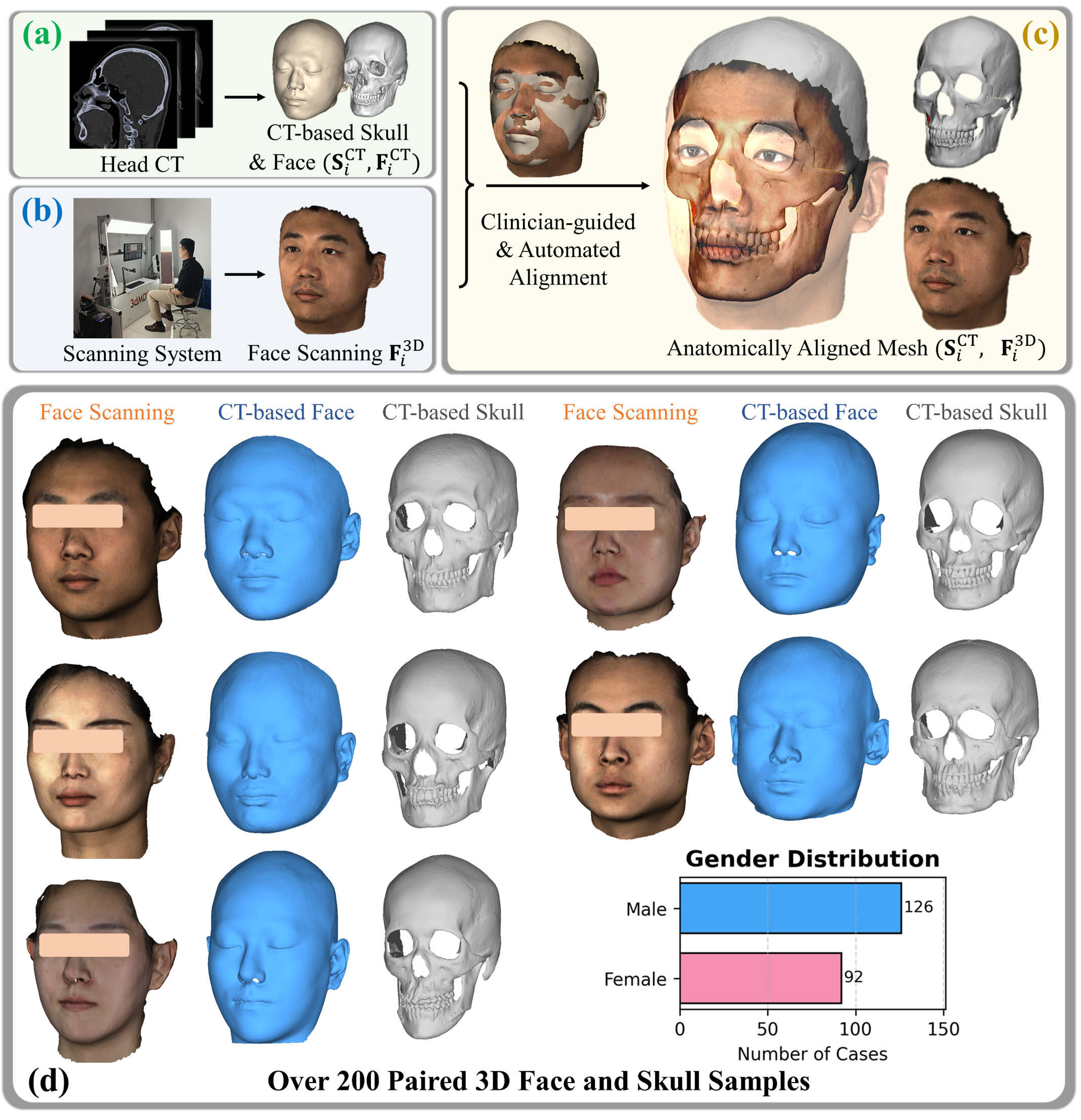}
\caption{(a) and (b): Overview of our face and skull data acquisition process. (c): Clinician-guided and automated alignment of CT and scanning data. (d): Visualization of paired 3D face and skull samples from our dataset.}
\label{preprocess}
\end{figure}

Since $\mathbf{F}_{i}^{\mathrm{CT}}$ lacks texture information and is typically captured with the subject's eyes closed, it is not ideal for directly constructing a morphable facial model. Therefore, we use $\mathbf{F}_{i}^{\mathrm{CT}}$ as an intermediate modality to anatomically align $\mathbf{F}_{i}^{\mathrm{3D}}$ with $\mathbf{S}_{i}^{\mathrm{CT}}$. Specifically, clinical experts first manually annotate anatomical landmarks on both $\mathbf{F}_{i}^{\mathrm{CT}}$ and $\mathbf{F}_{i}^{\mathrm{3D}}$ to perform an initial rigid alignment. This is followed by automated alignment using professional-grade medical modeling software, such as \textit{Materialise 3-matic Medical}. The registration results are then reviewed and refined by experienced clinicians to ensure anatomically accurate alignment between $\mathbf{F}_{i}^{\mathrm{3D}}$ and $\mathbf{S}_{i}^{\mathrm{CT}}$. An overview of this process is shown in Figure~\ref{preprocess}(c).

\begin{figure*}[t]
\centering
\includegraphics[width=1.0\textwidth]{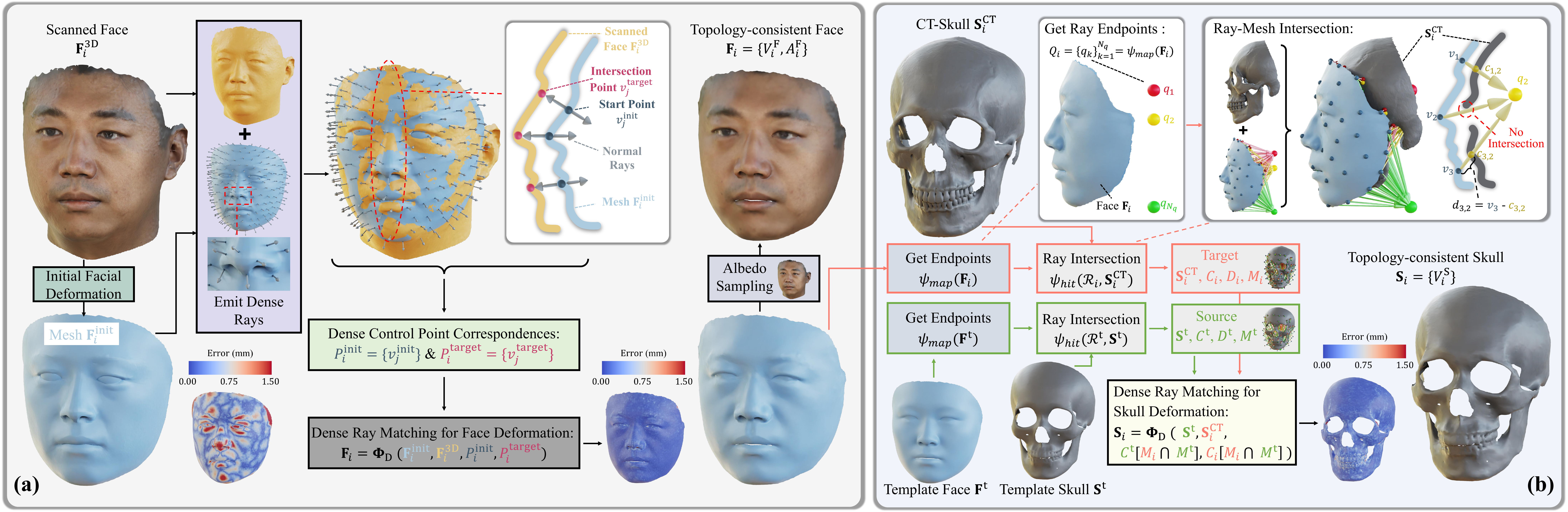}
\caption{Overview of our topology-consistent deformation pipeline. (a) Dense ray matching for face mesh deformation. (b) Dense ray matching for skull mesh deformation.}
\label{dense-matching}
\end{figure*}

\section{Modeling}

This section introduces a novel dense ray-matching deformation method to achieve topology-consistent face and skull registration and establish point-to-point anatomical tissue correspondences. Based on this, we present the Face-Skull Morphable Model and the Tissue Morphable Model, which together form the Bidirectional Face-Skull Morphable Model (BFSM), the foundation of Vision Tomography.

\subsection{Dense Ray Matching for Topology Registration}

Our goal is to deform a pair of fixed-topology face and skull templates (\( \mathbf{F}^{\mathrm{t}} \) and \( \mathbf{S}^{\mathrm{t}} \)) to match the anatomically aligned target data (\( \mathbf{F}_i^{\mathrm{3D}} \) and \( \mathbf{S}_i^{\mathrm{CT}} \)). This yields topology-consistent face \( \mathbf{F}_i \) and skull \( \mathbf{S}_i \), along with dense tissue correspondences $D_i$ between them.

We first perform non-rigid registration on the facial surface and then leverage its strong geometric priors to guide the skull topology registration, which subsequently yields dense tissue correspondences between the face and the skull. To support this, we introduce a novel dense ray-matching deformation method that establishes dense and anatomically consistent vertex correspondences for the non-rigid deformation module. Existing non-rigid deformation methods \cite{nicp_github, amberg2007optimal}, which are predominantly based on nearest-neighbor strategies, often rely heavily on accurately estimated control points to avoid convergence to local optima. Our dense ray-matching-based guidance focuses on providing thousands of semantically consistent control point pairs in the process of face-skull topology alignment, which significantly improves registration robustness.

\myparagraph{Initial Facial Deformation.}
For initialization, we first register the target face \( \mathbf{F}_i^{\mathrm{3D}} \) using BFM \cite{blanz1999morphable} with differentiable rendering \cite{Laine2020diffrast, ravi2020pytorch3d}. Specifically, the 3D facial scan \( \mathbf{F}_i^{\mathrm{3D}} \) is rendered to obtain its corresponding RGB image, depth map, and silhouette. We extract predicted facial attributes such as landmarks, depth, texture, and semantic part segmentation \cite{wang20243d} for supervision signals to iteratively optimize the coefficients of the BFM. The result of this optimization is a topology-consistent face mesh \( \mathbf{F}^{\mathrm{init}}_{i} \). Figure~\ref{dense-matching}(a) shows that \( \mathbf{F}^{\mathrm{init}}_{i} \) is coarsely aligned with \( \mathbf{F}_i^{\mathrm{3D}} \), with the alignment error visualized as a heatmap.

\myparagraph{Deformation Module \( \boldsymbol{\Phi}_\mathrm{D} \).} We define a non-rigid deformation module \( \boldsymbol{\Phi}_\mathrm{D} \) that deforms a source mesh \( \mathbf{X} \) to a target mesh \( \mathbf{X}' \), guided by control points. The module takes two 3D meshes \( \mathbf{X} \) and \( \mathbf{X}' \), and their corresponding sets of control points \( {P} \in \mathbb{R}^{k \cdot 3} \) and \( {P}'\in \mathbb{R}^{k \cdot 3} \) as input. Note that $\mathbf{X}$ and $\mathbf{X}'$ differ in their topological structures. \( \boldsymbol{\Phi}_\mathrm{D} \) outputs a registered mesh \( \mathbf{X}_{\mathrm{r}} \), which shares the same topology as $\mathbf{X}$ (\textit{i.e.}, the number and order of vertices are consistent, and the connectivity of the triangular faces is preserved). In the process, \( \mathbf{X}_{\mathrm{r}} \) converges towards the shape of \( \mathbf{X}' \):
\begin{equation}
\begin{aligned}
\begin{array}{l}
\mathbf{X}_{\mathrm{r}} = \boldsymbol{\Phi}_\mathrm{D}(\mathbf{X}, \mathbf{X}' , {P}, {P}')
\end{array}
\end{aligned},
\label{Basic_Deformation_Module}
\end{equation}
where the deformation module \( \boldsymbol{\Phi}_\mathrm{D} \) is defined as the solution to the following optimization objective:
\begin{equation}
\small
\begin{aligned}
\mathcal{L}_{\text{D}} &= \mathcal{L}_{\text{mesh}} + \lambda \mathcal{L}_{\text{smooth}} + \mu \mathcal{L}_{\text{P}}\\
 &=\sum_{i=1}^{N} \left\| \mathbf{T}({v}_i) - \Pi({v}_i, \mathbf{X}') \right\|^2  \\
 & \quad+ \lambda \sum_{i=1}^{N} \left\| \Delta \mathbf{T}(\mathbf{v}_i) \right\|^2  + \mu \left\| \mathbf{T}({P}) - {P}' \right\|^2,
\end{aligned}
\label{Basic_Deformation_Module_Loss}
\end{equation}
where the deformation \( \mathbf{T}(\cdot) \) is optimized as a vertex affine field using the AMSGrad \cite{reddi2019convergence} optimizer within a PyTorch \cite{paszke2019pytorch} framework \cite{nicp_github, amberg2007optimal}. $\{v_i\}_{i=1}^{N}$ are the vetices in mesh \(\mathbf{X}\). $\mathcal{L}_{\text{mesh}}$ encourages mesh alignment between the source and the target via nearest-neighbor matching, \( \Pi({v}_i, \mathbf{X}') \) finds the closest point on the target mesh \( \mathbf{X}' \) to the vertex \( {v}_i \). The smoothness term \( \mathcal{L}_{\text{smooth}} \) enforces Laplacian regularity \cite{amberg2007optimal} to ensure locally smooth deformations, where \( \Delta \) is the mesh Laplacian operator. The control point term $\mathcal{L}_{\text{P}}$ ensures anatomical consistency at landmark locations.

\( \boldsymbol{\Phi}_\mathrm{D} \) relies on accurate control point correspondences to avoid local optima and achieve topology-consistent registration, practical scenarios often provide only a few dozen sparse landmarks on the face or skull. Due to the sparsity of control points, even minor inaccuracies can lead to significant registration errors. We therefore propose a dense ray-matching-based guidance to extract tens of thousands of anatomically plausible correspondences, significantly improving the effectiveness and robustnes of \( \boldsymbol{\Phi}_\mathrm{D} \).

\myparagraph{Dense Ray Matching for Face Deformation.} As shown in Figure~\ref{dense-matching}(a), starting from the surface of \( \mathbf{F}_i^{\mathrm{init}} \), we emit two rays from each vertex \( {v}_j^{\mathrm{init}} \in \mathbf{F}_i^{\mathrm{init}} \), following the directions of the positive and negative vertex normals. These rays, visualized in Figure~\ref{dense-matching}(a), intersect the target scan \( \mathbf{F}_i^{\mathrm{3D}} \) with each intersection point denoted as \( {v}_j^{\mathrm{target}} \). Ray origins and intersections construct two dense point sets, ${P}_i^{\mathrm{init}} = \{ {v}_j^{\mathrm{init}} \}$ and ${P}_i^{\mathrm{target}}  = \{ {v}_j^{\mathrm{target}} \}$, establishing dense control point correspondences between \( \mathbf{F}_i^{\mathrm{init}} \) and \( \mathbf{F}_i^{\mathrm{3D}} \). These correspondences are used as supervision signals for the module \( \boldsymbol{\Phi}_\mathrm{D} \):
\begin{equation}
\begin{aligned}
\begin{array}{l}
 \mathbf{F}_i  = \boldsymbol{\Phi}_\mathrm{D}(\mathbf{F}_i^{\mathrm{init}}, \mathbf{F}_i^{\mathrm{3D}} , {P}_i^{\mathrm{init}}, {P}_i^{\mathrm{target}})
\end{array}
\end{aligned},
\label{face_Deformation}
\end{equation}
where we obtain the topology-consistent face mesh \( \mathbf{F}_i = \{ {V}_i^\mathrm{F}, {A}_i^\mathrm{F} \} \) with fixed triangle-faces. \( {V}_i^\mathrm{F} \in \mathbb{R}^{N_\mathrm{F} \cdot 3} \) represents the vertices and \( {A}_i^\mathrm{F} \) denotes the per-vertex albedo. The alignment accuracy between \( \mathbf{F}_i \) and the input scan \( \mathbf{F}_i^{\mathrm{3D}} \) is visualized as an error heatmap in Figure~\ref{dense-matching}(a), showing that \( \mathbf{F}_i \) closely matches the geometry of \( \mathbf{F}_i^{\mathrm{3D}} \). In rare cases where the normal rays fail to intersect the target facial mesh $ \mathbf{F}_i^{\mathrm{3D}} $, the nearest point to \( {v}_j^{\mathrm{init}} \) on \( \mathbf{F}_i^{\mathrm{3D}} \) is used as \( {v}_j^{\mathrm{target}} \).

\myparagraph{Dense Ray Matching for Skull Deformation.} As shown in Figure~\ref{dense-matching}(b), based on the face deformation step, we have a pair of anatomically aligned meshes \( \mathbf{F}_i \) and \( \mathbf{S}_i^{\mathrm{CT}} \). Our goal is to deform and register the skull template \( \mathbf{S}^{\mathrm{t}} \) from a pair of fixed-topology face and skull templates (\( \mathbf{F}^{\mathrm{t}} \) and \( \mathbf{S}^{\mathrm{t}} \)) to align with \( \mathbf{S}_i^{\mathrm{CT}} \). Note that \( \mathbf{F}_i \) and \( \mathbf{F}^{\mathrm{t}} \) share the same topology, whereas \( \mathbf{S}_i^{\mathrm{CT}} \) and \( \mathbf{S}^{\mathrm{t}} \) do not. Due to the more complex geometry of the skull, including holes and other anatomical structures, we adopt a dense ray matching strategy to leverage the structural relationship between the face and the skull. This allows us to obtain the control points needed to deform the skull template \( \mathbf{S}^{\mathrm{t}} \) to the target CT skull \( \mathbf{S}^{\mathrm{CT}} \) in the deformation module \( \boldsymbol{\Phi}_\mathrm{D} \) and establish dense tissue correspondences between the face and the skull.

Given two anatomically aligned pairs (\( \mathbf{F}_i \), \( \mathbf{S}_i^{\mathrm{CT}} \)) and (\( \mathbf{F}^{\mathrm{t}} \), \( \mathbf{S}^{\mathrm{t}} \)), we first construct a fixed mapping function \( \psi_\mathrm{map}(\cdot) \), which maps facial landmarks on the face mesh to a series of spatially meaningful 3D points. These points are located in or near the occipital region of the head and serve as the endpoints of dense rays used for skull deformation. Specifically, \( \psi_\mathrm{map}(\cdot) \) applies a fixed transformation to a subset of the facial landmarks (such as the chin tip, nose tip, and the midpoint of the two outer eye corners), adjusting their z-coordinates based on the overall depth of the face to align anatomically with the occipital region of the head. We apply \( \psi_\mathrm{map}(\cdot) \) to \( \mathbf{F}_i \) and \( \mathbf{F}^{\mathrm{t}} \), respectively, obtaining two sets of endpoints \( Q_i = \{ {q}_k \}_{k=1}^{N_q} \) and \( Q^{\mathrm{t}} = \{ {q}_k^{\mathrm{t}} \}_{k=1}^{N_q} \). For the subject $i$ and template face meshes (\( \mathbf{F}_i \) and \( \mathbf{F}^{\mathrm{t}}\)), their vertex sets are denoted as \( {V}_i^{\mathrm{F}} = \{ {v}_j \}_{j=1}^{N_\mathrm{F}} \in \mathbf{F}_i \) and \( {V}^{\mathrm{t}} = \{ {v}_j^{\mathrm{t}} \}_{j=1}^{N_\mathrm{F}} \in \mathbf{F}^{\mathrm{t}} \), respectively. For each mesh, we construct a set of rays \( \mathcal{R}_i = \{ \mathbf{r}_{j,k} \} \) and \( \mathcal{R}^{\mathrm{t}} = \{ \mathbf{r}_{j,k}^{\mathrm{t}} \} \), where each ray originates from a facial vertex \( {v}_j \) or \( {v}_j^{\mathrm{t}} \) and points toward a endpoint \( {q}_k \in Q_i \) or \( {q}_k^{\mathrm{t}} \in Q^{\mathrm{t}} \), as shown in  Figure~\ref{dense-matching}(b).

Since each ray in \( \mathcal{R}_i \) and \( \mathcal{R}^{\mathrm{t}} \) has an ordered pair of origin and end points, the corresponding intersection points with the skull meshes \( \mathbf{S}_i^{\mathrm{CT}} \) and \( \mathbf{S}^{\mathrm{t}} \) are also ordered accordingly. We introduce a ray-mesh intersection function, denoted as \( \psi_{\mathrm{hit}}(\cdot) \), to compute these intersections on the skulls:
\begin{equation}
\begin{aligned}
C_i, D_i, M_i &= \psi_{\mathrm{hit}}(\mathcal{R}_i, \mathbf{S}_i^{\mathrm{CT}})\\C^{\mathrm{t}}, D^{\mathrm{t}}, M^{\mathrm{t}} &= \psi_{\mathrm{hit}}(\mathcal{R}^{\mathrm{t}}, \mathbf{S}^{\mathrm{t}})
\end{aligned}
\label{dense_skull_ray},
\end{equation}
where \( C_i = \{c_{j,k}\} \in \mathbb{R}^{(N_\mathrm{F} \cdot N_q) \cdot 3} \) and \( C^{\mathrm{t}} = \{{c_{j,k}}^{\mathrm{t}}\} \in \mathbb{R}^{(N_\mathrm{F} \cdot N_q) \cdot 3} \)  represent the intersection points on skull. \( D_i = \{d_{j,k}\} \in \mathbb{R}^{(N_\mathrm{F} \cdot N_q) \cdot 3} \) and \( D^{\mathrm{t}} \in \mathbb{R}^{(N_\mathrm{F} \cdot N_q) \cdot 3} \) denote the direction vectors from the skull to the face with distance, where \(d_{j,k} = v_j - c_{j,k}\), \(v_j \in \mathbf{F}_i\). \( D_i \) is a dense estimation of the tissue thickness between the face and the skull, which is later used for constructing the Tissue Morphable Model. The masks \( M_i, M^{\mathrm{t}} \in \{0,1\}^{N_\mathrm{F} \cdot N_q} \) indicate the validity of each intersection. Due to anatomical features such as holes and discontinuities in the skull geometry, some rays \( \mathbf{r}_{j,k} \in \mathcal{R}_i \) or \( \mathbf{r}_{j,k}^{\mathrm{t}} \in \mathcal{R}^{\mathrm{t}} \) may not produce valid intersections. In such cases, \( \psi_{\mathrm{hit}}(\cdot) \) assigns a sentinel value and marks these entries as invalid in the corresponding masks \( M_i \) and \( M^{\mathrm{t}} \).

\begin{figure}[t]
\centering
\includegraphics[width=1.0\columnwidth]{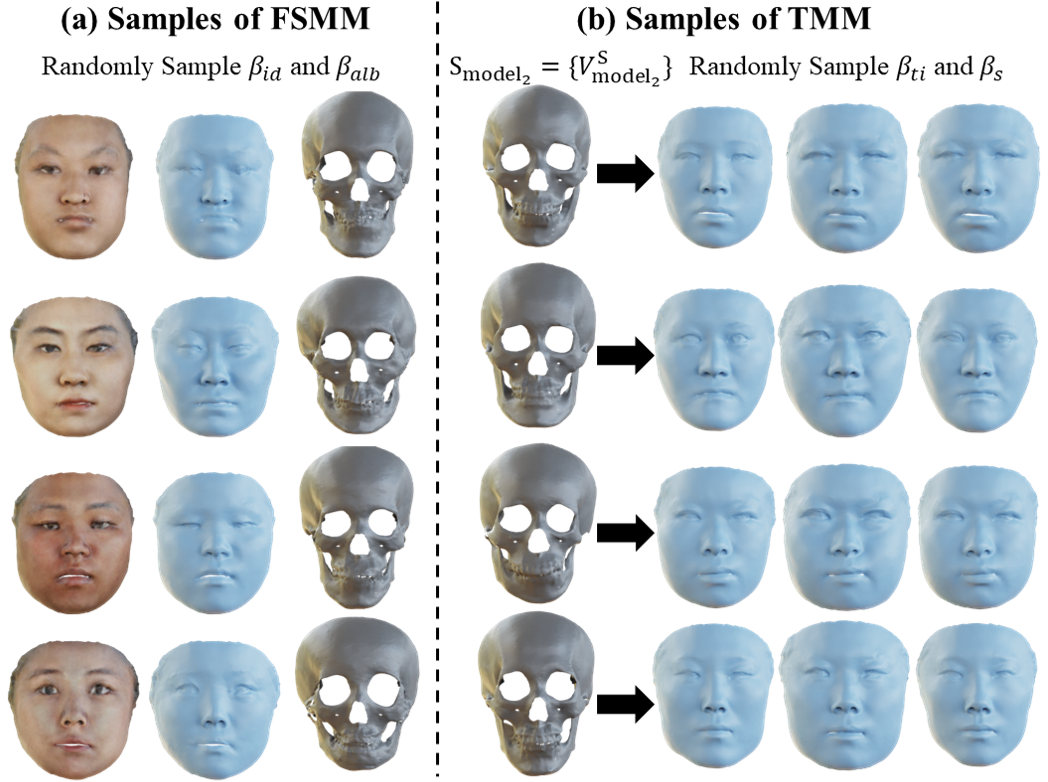}
\caption{Anatomically aligned face-skull pairs generated by randomly sampling \((\beta_{{id}}, \beta_{{alb}})\) from FSMM and \((\beta_{{ti}}, \beta_s)\) from TMM, respectively.}
\label{Visualization}
\end{figure}

At this stage, we have acquired the dense skull control points required by the deformation module \( \boldsymbol{\Phi}_\mathrm{D} \) for the pairs \( (\mathbf{F}_i, \mathbf{S}_i^{\mathrm{CT}}) \) and \( (\mathbf{F}^{\mathrm{t}}, \mathbf{S}^{\mathrm{t}}) \), enabling the deformation of the skull template \( \mathbf{S}^{\mathrm{t}} \) to the target skull \( \mathbf{S}_i^{\mathrm{CT}} \):
\begin{equation}
\begin{aligned}
\begin{array}{l}
 \mathbf{S}_i  = \boldsymbol{\Phi}_\mathrm{D}(\mathbf{S}^{\mathrm{t}}, \mathbf{S}_i^{\mathrm{CT}},  C^{\mathrm{t}} [M_i \cap M^{\mathrm{t}}], C_i[M_i \cap M^{\mathrm{t}}])
\end{array}
\end{aligned},
\label{skull_Deformation}
\end{equation}
where we obtain the topology-consistent skull mesh \( \mathbf{S}_i = \{ {V}_i^\mathrm{S} \} \) with fixed triangle-faces. \( {V}_i^\mathrm{S} \in \mathbb{R}^{N_\mathrm{S} \cdot 3} \). As shown in Figure~\ref{dense-matching}(b), the template pair is fixed. For each subject \(i\), we apply the non-rigid deformation process based on dense ray matching as defined in Equations~\ref{face_Deformation} and~\ref{skull_Deformation}. This results in a collection of \(N_{\text{sample}}\) topologically aligned face-skull mesh pairs \(\{ \mathbf{F}_i, \mathbf{S}_i \}_{i=1}^{N_{\text{sample}}}\), along with corresponding dense tissue thickness \(\{ D_i \}_{i=1}^{N_{\text{sample}}}\) between face and skull. The impact of ray origin and endpoint numbers (\(N_O\) and \(N_q\)) on registration accuracy is analyzed in the ablation study.

To support Tissue Morphable Model in the next section,  
we construct a fixed index mapping \( \mathcal{I}_{\mathrm{t}} \in \mathbb{R}^{N_\mathrm{F} \cdot N_q} \) between the template skull vertices \( V^{\mathrm{t}}_\mathrm{S} \in \mathbb{R}^{N_\mathrm{S} \cdot 3} \) and the control points \( C^{\mathrm{t}} \in \mathbb{R}^{(N_\mathrm{F} \cdot N_q) \times 3} \),  
where each index is determined by nearest-neighbor search from \( C^{\mathrm{t}} \) to \( V^{\mathrm{t}}_\mathrm{S} \). \( \mathcal{I}_{\mathrm{t}} \) is later used to locate the origin vertices for the tissue thickness vectors on any given skull mesh.

\begin{figure}[t]
\centering
\includegraphics[width=1.0\columnwidth]{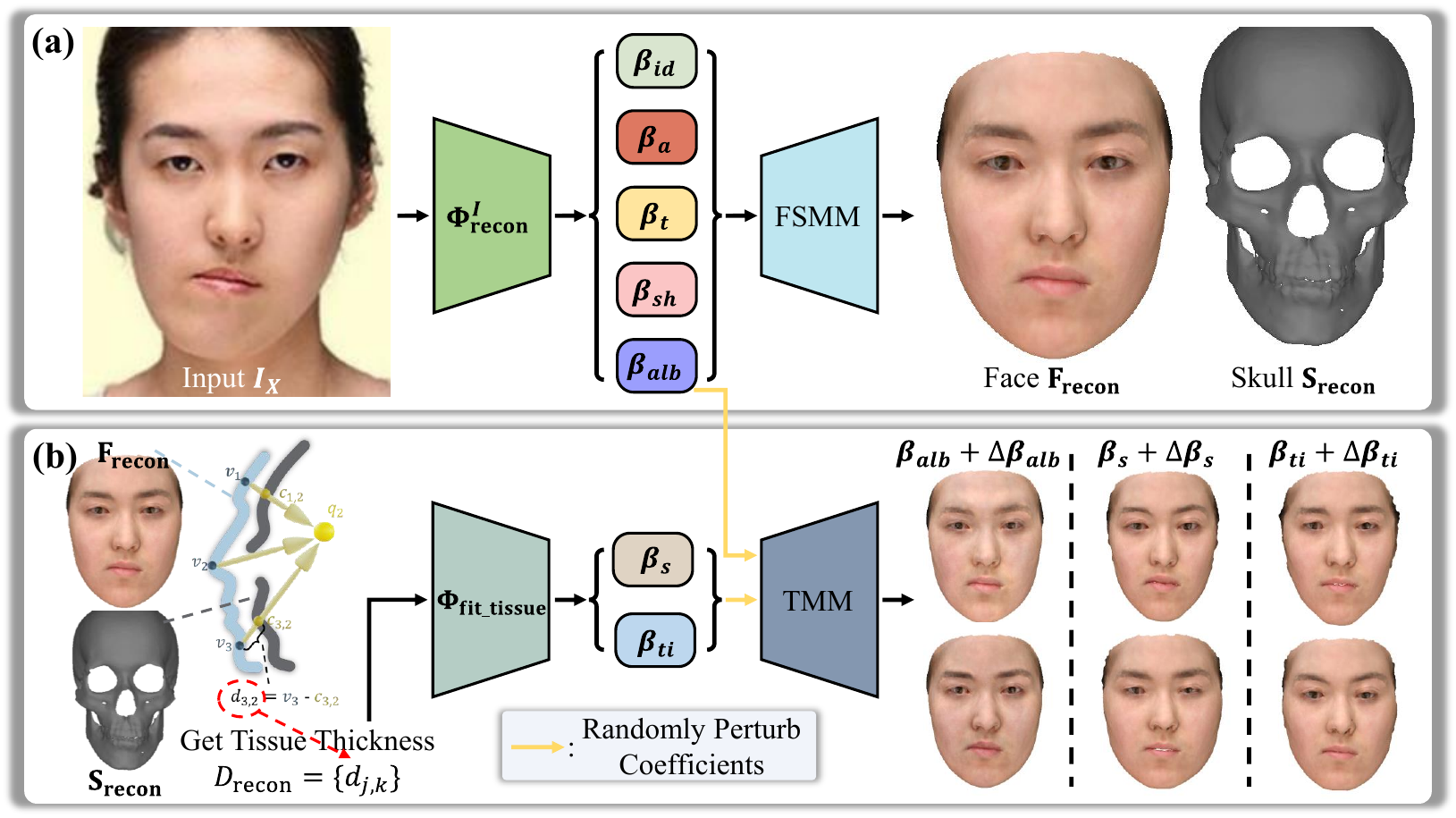}
\caption{(a) Given an input face image \( I_X \), VT predicts the coefficients of the FSMM to reconstruct an anatomically aligned 3D face-skull pair.
(b) Based on the reconstructed face-skull pair, the tissue thickness is estimated and the TMM coefficients are fitted, enabling facial variation by perturbing the TMM coefficients for a given skull.}
\label{Reconstruct_Paired_Face_and_Skull_from_Image}
\end{figure}

\subsection{Bidirectional Face-Skull Morphable Model (BFSM)}

To enhance robustness, each subject \( i\) is horizontally flipped, yielding \( N_{\text{sample}} = 161 \cdot 2 = 322 \) samples.

\myparagraph{Face-Skull Morphable Model.} 
Based on prior work on morphable models~\cite{wood2021fake, blanz1999morphable,wang2025srmhair}, and Principal Component Analysis (PCA)~\cite{Sirovich:87, jolliffe2016principal}, we construct a {3D Face--Skull Morphable Model (FSMM)}. FSMM enables the generation of a new pair of anatomically aligned face and skull from a {shared set of coefficients}:
\begin{equation}
\begin{aligned}
V_\mathrm{model_1}^{\mathrm{FS}} &= \left[ V_\mathrm{model_1}^\mathrm{F} \,\middle\|\, V_\mathrm{model_1}^\mathrm{S} \right] \\ &= \boldsymbol{R}({{\beta} _{a}})({\overline V}  + {{\beta} _{id}}{\boldsymbol{B}_{id}}) + {{\beta} _{t}}\\
A_\mathrm{model_1}^\mathrm{F} &= {\overline A}  + {{\beta} _{alb}}{\boldsymbol{B}_{alb}}
\end{aligned}\quad,
\label{FSMM-equ}
\end{equation}
where \( {\beta}_{id} \in \mathbb{R}^{n_{id}} \), \( {\beta}_{alb} \in \mathbb{R}^{n_{alb}} \), \( {\beta}_{a} \in \mathbb{R}^{3} \), and \( {\beta}_{t} \in \mathbb{R}^{3} \) are the identity, albedo, rotation, and translation coefficients, respectively. \( \boldsymbol{R}({\beta}_{a}) \in \mathbb{R}^{3 \cdot 3} \) is the rotation matrix corresponding to angles \( {\beta}_{a} \in \mathbb{R}^{3} \). \( \overline{V} \in \mathbb{R}^{(N_{\mathrm{F}} + N_{\mathrm{S}}) \cdot 3} \) and \( \overline{A} \in \mathbb{R}^{N_{\mathrm{F}} \cdot 3} \) denote the mean shape and mean albedo, computed from the face-skull joint vertex set \( \left[ V_{i}^\mathrm{F} \,\middle\|\, V_{i}^\mathrm{S} \right]_{i=1}^{N_{\text{sample}}} \) and the facial texture set \( \{ A_i^{\mathrm{F}} \}_{i=1}^{N_{\text{sample}}} \), respectively. \( \boldsymbol{B}_{id} \) and \( \boldsymbol{B}_{alb} \) are obtained by applying PCA to the concatenated vertex pairs \( \left[ V_{i}^\mathrm{F} \,\middle\|\, V_{i}^\mathrm{S} \right]_{i=1}^{N_{\text{sample}}} \) and the facial albedo set \( \{ A_i^{\mathrm{F}} \}_{i=1}^{N_{\text{sample}}} \), respectively. The operator \( \left[ \cdot \,\middle\|\, \cdot \right] \) denotes vertex concatenation.

\begin{figure*}[!htbp]
\centering
\includegraphics[width=0.93\textwidth]{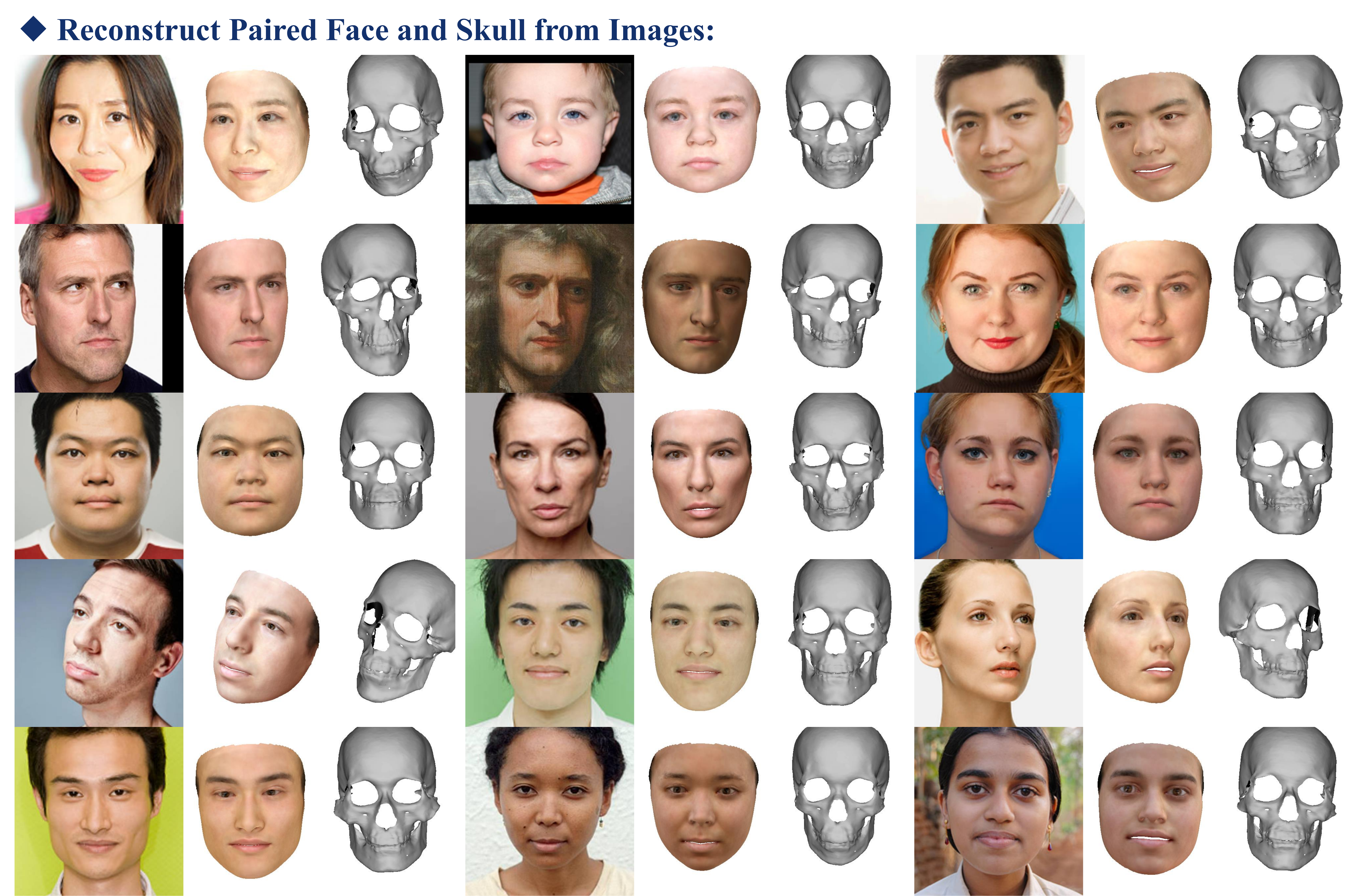}
\caption{Paired 3D face and skull reconstruction from images.}
\label{recon_res1}
\end{figure*}

\begin{figure*}[!htbp]
\centering
\includegraphics[width=0.93\textwidth]{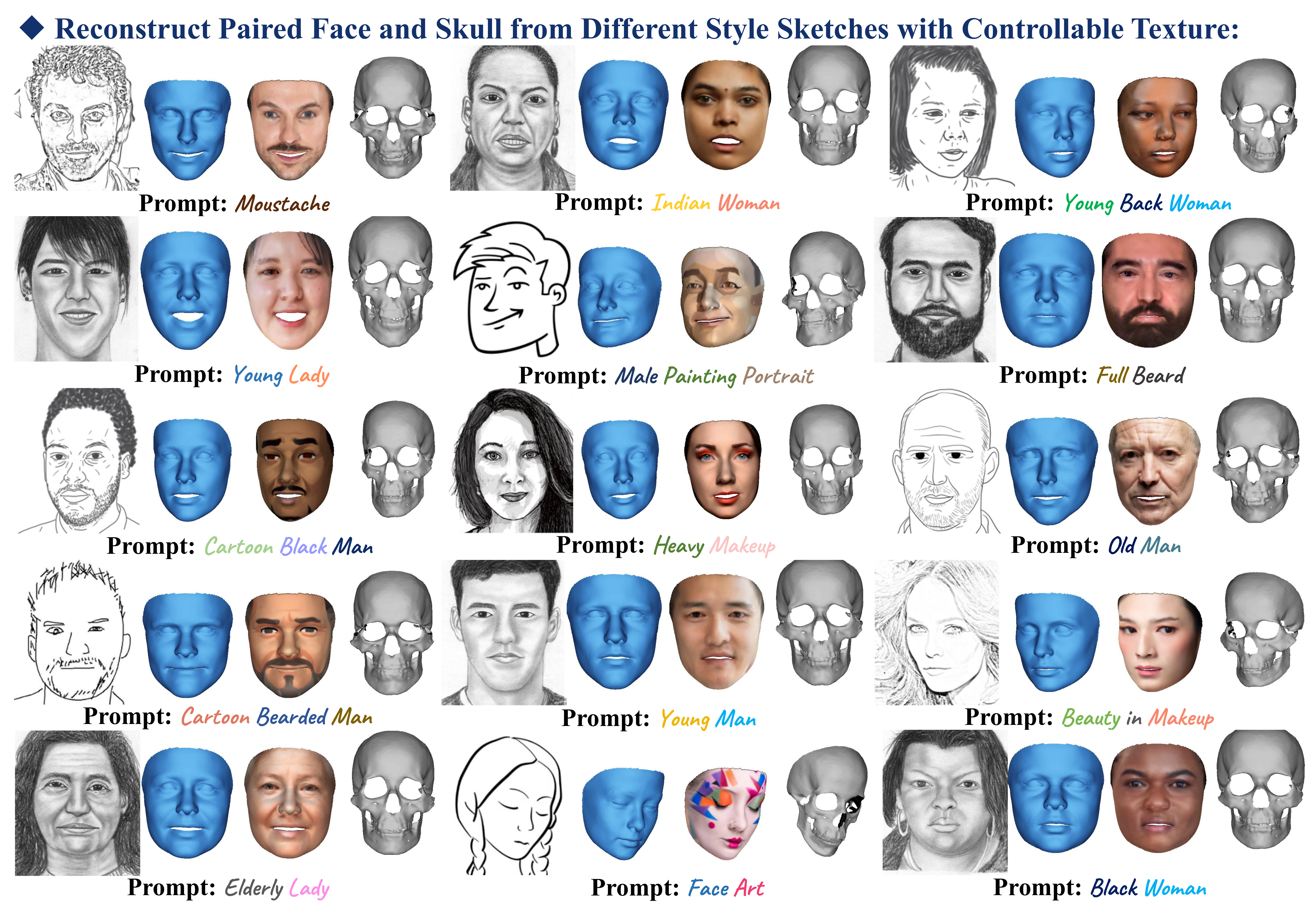}
\caption{Paired 3D face and skull reconstruction from different style sketches with controllable facial texture.}
\label{recon_res2}
\end{figure*}

\myparagraph{Tissue Morphable Model.} In real-world scenarios, a single skull may correspond to multiple facial appearances due to factors like age or body fat. To capture this one-to-many relationship, we construct a Tissue Morphable Model (TMM) based on tissue thickness vectors \( \{ D_i \}_{i=1}^{N_{\text{sample}}} \). The tissue depth from skull to face can be generated by:
\begin{equation}
{D}_{\mathrm{model_2}} = \left( \overline{{D}} + {{\beta} _{ti}}{\boldsymbol{B}_{ti}} \right) \cdot \beta_s
\label{TMM}\quad,
\end{equation}
where \( \overline{D} \in \mathbb{R}^{(N_{\mathrm{F}} \cdot N_q) \times 3} \) denotes the mean tissue thickness vector,  \( \boldsymbol{B}_{ti}  \) represents the PCA basis of tissue thickness \( \{ D_i \}_{i=1}^{N_{\text{sample}}} \),  
\( \beta_{ti} \in \mathbb{R}^{n_{ti}} \) is the individual-specific tissue coefficients, and \( \beta_s \in \mathbb{R} \) is a global scalar controlling overall tissue thickness scaling. Given a skull mesh \( \mathbf{S}_\mathrm{model_2} = \{ V_\mathrm{model_2}^\mathrm{S} \} \) with the same fixed topology as \( \mathbf{S}^{\mathrm{t}} \), a new set of facial surface points can be generated as:
\begin{equation}
\begin{aligned}
V'_{\mathrm{model_2}} &= V_\mathrm{model_2}^\mathrm{S}\left[\mathcal{I}_{\mathrm{t}} \right] + {D}_{\mathrm{model_2}} \quad,\\
 V_\mathrm{model_2}^\mathrm{F} &= \boldsymbol{\Phi}_\mathrm{fit}(V'_{\mathrm{model_2}})
\label{fit_TMM}\quad,
\end{aligned}
\end{equation}
where \( \mathcal{I}_{\mathrm{t}} \in \mathbb{R}^{(N_{\mathrm{F}} \cdot N_q)} \) is the precomputed index map that associates each tissue thickness vector to a vertex on the given skull \( \mathbf{S}_\mathrm{model_2} \), resulting \(V_\mathrm{model_2}^\mathrm{S}\left[\mathcal{I}_{\mathrm{t}} \right]  \in \mathbb{R}^{(N_{\mathrm{F}} \cdot N_q) \times 3} \) and initial face vertices \(V'_{\mathrm{model_2}} \in \mathbb{R}^{(N_{\mathrm{F}} \cdot N_q) \times 3}\), and \( \boldsymbol{\Phi}_{\mathrm{fit}}(\cdot) \) denotes a face mesh reconstruction function that fits the facial surface points \( V'_{\mathrm{model_2}} \) onto a topology-consistent facial mesh. If the given skull mesh does not share the same topology as \( \mathbf{S}^{\mathrm{t}} \), a similar deformation strategy can be applied to obtain a topology-consistent representation, which we omit here for brevity. 
Note that the facial pose in TMM is inherently determined by the input skull \( \mathbf{S}_\mathrm{model_2} \) and Equation~\ref{fit_TMM} does not require additional pose coefficients.

 \( \boldsymbol{\Phi}_{\mathrm{fit}}(\cdot) \) could be implemented through a deformation process similar to Equations~\ref{face_Deformation} and~\ref{skull_Deformation} or using FSMM. For example, given the tissue thickness vector $D_{\mathrm{model_2}}$ and the skull vertices $V^\mathrm{S}_{\mathrm{model_2}}$, $\boldsymbol{\Phi}_{\mathrm{fit}}(\cdot)$ fits the facial surface points $V'_{\mathrm{model_2}}$ onto a topology-consistent mesh. This is achieved via a multi-stage optimization initialized with FSMM identity and pose coefficients. The process involves rigid alignment stages, optimizing rotation, scale, translation, \textit{etc.}, and optionally texture, followed by non-rigid refinement stages incorporating identity coefficients to capture personalized geometry. Learning rates are progressively reduced across stages to ensure stable convergence.

\myparagraph{Visualization.} Figure~\ref{Visualization}(a) and Figure~\ref{Visualization}(b) illustrate anatomically aligned face-skull pairs generated by randomly sampling \((\beta_{{id}}, \beta_{{alb}})\) from FSMM and \((\beta_{{ti}}, \beta_s)\) from TMM, respectively.

\section{Applications}
\label{4Reconstruction}

This section introduces the applications of Vision Tomography, including paired face-skull reconstruction from visual inputs, joint editing of face and skull geometry for surgical planning, \textit{etc}.

\subsection{Face and Skull Reconstruction from Various Visual Inputs}

Reconstructing paired 3D face and skull from visual data is important for telemedicine, missing people search, medical education, and forensics. Additionally, in clinical practice, some aesthetic treatments require multiple surgeries for correction, often necessitating repeated CT scans to capture the patient's 3D face and skull geometry and shape changes during the treatment process. The frequent use of CT scans exposes patients to potentially harmful radiation, which cannot be ignored. To address these issues and enable access to 3D face and skull data in economically underdeveloped regions without relying on expensive CT equipment, we propose a portable, non-invasive framework capable of reconstructing paired 3D face-skull geometry from multi-modal inputs such as images, sketches, and text descriptions. The process is as follows.

We assume the visual facial input is denoted as \( X \), which can take one of two forms: RGB face images \( I_X \) and sketches \( S_X \). Following the method in \cite{wang2024s2td}, we assume that when the input modality is a sketch \( S_X \), it may optionally include a corresponding face texture description \( T_X^S \), as sketches alone lack texture information of the face. The FSMM we developed enables the construction of anatomically aligned face-skull pairs from shared coefficients (\(\beta=\{{\beta}_{id}, {\beta}_{a},{\beta}_{t},{\beta}_{alb}\}\)). The key focus of the reconstruction framework is to explore how to construct \( \boldsymbol{\Phi}_{\mathrm{recon}} = \{\boldsymbol{\Phi}_{\mathrm{recon}}^I,\boldsymbol{\Phi}_{\mathrm{recon}}^S \} \) to predict \(\beta\) from visual data \( X = \{ I_X , S_X\} \). 

For the image input \( I_X \), we leverage existing monocular 3D face reconstruction methods~\cite{wang2024s2td, wang20243d, DECA:Siggraph2021, deng2019accurate} to supervise the face component \( V_\mathrm{model_1}^\mathrm{F} \) in Equation~\ref{FSMM-equ} using only 2D facial image training data. Specifically, we train a backbone \( \boldsymbol{\Phi}_\mathrm{recon}^I \) to predict the coefficients of the FSMM, thereby enabling the reconstruction of an anatomically aligned face-skull pair (\( \mathbf{F}_\mathrm{recon} \), \( \mathbf{S}_\mathrm{recon} \)) from a single input image \( I_X \). For sketch input \( S_X \) and its corresponding face texture description \( T_X^S \), we follow \cite{wang2024s2td} to train a \( \boldsymbol{\Phi}_\mathrm{recon}^S \), enabling the reconstruction of an anatomically aligned face-skull pair (\( \mathbf{F}_\mathrm{recon} \), \( \mathbf{S}_\mathrm{recon} \)). During the training of \( \boldsymbol{\Phi}_\mathrm{recon}^S \), we first transform existing abundant face image datasets into sketches of various styles \cite{bradski2000opencv, simo2018mastering, simo2016learning} to serve as input to \( \boldsymbol{\Phi}_\mathrm{recon}^S \), with the original RGB images providing supervision. The face texture description \( T_X^S \) is used to obtain textures in UV space from a given texture library \cite{wang2024s2td}, which are then seamlessly mapped onto the predicted 3D face geometry. It also could be used for generating face images using ControlNet \cite{zhang2023adding}, followed by extracting UV textures.

As illustrated in Figure~\ref{Reconstruct_Paired_Face_and_Skull_from_Image}(a), we take a face image \( I_X \) as an example to illustrate the process of predicting the coefficients of the FSMM, thereby enabling the reconstruction of an anatomically aligned face-skull pair (\( \mathbf{F}_\mathrm{recon} \), \( \mathbf{S}_\mathrm{recon} \)). In Figure~\ref{Reconstruct_Paired_Face_and_Skull_from_Image}(a), \( {\beta}_{sh} \) denotes the Spherical Harmonics (SH) \cite{ramamoorthi2001efficient} coefficients used for facial appearance modeling. Figure~\ref{recon_res1} and Figure~\ref{recon_res2} provides more reconstruction results of our method.

Figure~\ref{Reconstruct_Paired_Face_and_Skull_from_Image}(b) shows simple facial attractiveness adjustments based on the predicted face and skull geometry. Given the 3D face-skull pair \( \mathbf{F}_\mathrm{recon} \) and \( \mathbf{S}_\mathrm{recon} \), the tissue thickness vector \( D_{\mathrm{recon}} \) can be estimated using Equation~\ref{dense_skull_ray}. Based on this, we design a gradient-based optimization framework \( \boldsymbol{\Phi}_\mathrm{fit\_tissue} \) to fit the TMM coefficients \((\beta_{ti}, \beta_{s})\) corresponding to \( D_{\mathrm{recon}} \):
\begin{equation}
\beta_{ti}, \beta_{s}= \boldsymbol{\Phi}_\mathrm{fit\_tissue} \left({D}_{\mathrm{recon}}  \right)
\label{fittissue}\quad,
\end{equation}
which enables us to keep the skull fixed while sampling perturbed coefficients to generate diverse facial variations.

\begin{table*}[t]
\centering
\caption{Quantitative comparison of 3D face-skull reconstruction from images. The best is highlighted in bold. Distance errors are measured in millimeters, and a recall threshold of $3mm$ is used.}
\setlength{\tabcolsep}{0.9mm} 
\begin{tabular}{c|cccccc|cccccc}
\toprule[1pt]
\multirow{3}{*}{Method} & \multicolumn{6}{c|}{\textit{Frontal-view Image as Input}}                                              & \multicolumn{6}{c}{\textit{Side-view Image as Input}}                                                  \\ \cline{2-13} 
                        & \multicolumn{3}{c|}{\textit{Face}}                        & \multicolumn{3}{c|}{\textit{Skull}}  & \multicolumn{3}{c|}{\textit{Face}}                        & \multicolumn{3}{c}{Skull}   \\ 
                        & RMSE $\downarrow$   & Recall$\uparrow$  & \multicolumn{1}{c|}{Chamfer $\downarrow$} & RMSE $\downarrow$   & Recall $\uparrow$  & Chamfer $\downarrow$ & RMSE $\downarrow$   & Recall $\uparrow$  & \multicolumn{1}{c|}{Chamfer $\downarrow$} & RMSE $\downarrow$   & Recall $\uparrow$  & Chamfer $\downarrow$ \\ \midrule[0.5pt]
SMIRK+FSMM                   & 2.681 & 0.622 & \multicolumn{1}{c|}{6.591} & 2.444 & 0.688 & 5.445 & 3.007 & 0.555 & \multicolumn{1}{c|}{7.099} & 2.604 & 0.661 & 5.871 \\
DECA+FSMM                   & 2.674 & 0.614 & \multicolumn{1}{c|}{6.755} & 2.380 & 0.694 & 5.308 & 3.218 & 0.519 & \multicolumn{1}{c|}{7.616} & 2.637 & 0.658 & 5.915 \\
Deep3D+FSMM                   & 1.908 & 0.788 & \multicolumn{1}{c|}{3.994} & 2.044 & 0.758 & 4.395 & 2.311 & 0.712 & \multicolumn{1}{c|}{4.768} & 2.179 & 0.732 & 4.762 \\
3DDFA+FSMM                   & 2.392 & 0.675 & \multicolumn{1}{c|}{5.00} & 2.436 & 0.687 & 5.361 & 2.649 & 0.631 & \multicolumn{1}{c|}{5.487} & 2.768 & 0.631 & 6.104 \\
Ours                   & \textbf{1.864} & \textbf{0.801} & \multicolumn{1}{c|}{\textbf{3.905}} & \textbf{2.024} & \textbf{0.763} & \textbf{4.358} & \textbf{2.282} & \textbf{0.717} & \multicolumn{1}{c|}{\textbf{4.706}} & \textbf{2.171} & \textbf{0.733} & \textbf{4.748} \\\bottomrule[1pt]\end{tabular}
\label{Inference-based}
\end{table*}

\begin{table}
\centering
\caption{Quantitative comparison of the upper bounds of different face model representation capabilities. Distance errors are measured in millimeters, and a recall threshold of $1mm$ is used.}
\label{optimize-based}
\setlength{\tabcolsep}{2.82mm} 
\begin{tabular}{c|cc|ccc}
\toprule[1pt]
Method                 & $n_{id}$ & $n_{exp}$ & RMSE $\downarrow$  & Recall $\uparrow$ & Chamfer $\downarrow$ \\ \midrule[0.5pt]
\multirow{2}{*}{FLAME} & 300 & 0    & 0.520 & 0.910 & 7.539  \\
                       & 300 & 100  & 0.497 & 0.918 & 7.527   \\ \midrule[0.5pt]
\multirow{2}{*}{BFM}   & 199 & 0    & 0.766 & 0.726 & 1.881  \\
                       & 199 & 79$^\dagger$   & 0.616 & 0.814 & 1.625 \\ \midrule[0.5pt]
\multirow{4}{*}{Ours}  & 199 & 0    & 0.448 & 0.909 & 1.275  \\
                       & 300 & 0    & 0.421 & 0.923 & 1.230  \\
                       & 322 & 0    & 0.419 & 0.924 & 1.226  \\
                       & 322 & 79$^\dagger$   & \textbf{0.340} & \textbf{0.960} & \textbf{1.049}  \\ \bottomrule[1pt]
\end{tabular}
\begin{tablenotes}
\small
\item $^\dagger$ FaceWarehouse is used as the expression basis for BFM to enhance its representational capacity for fair comparison.
\end{tablenotes}
\end{table}

\subsection{Joint Editing of Face and Skull Geometry for Surgical Planning Prediction}

In clinical surgeries such as aesthetic plastic surgery and craniofacial deformity correction, surgeons typically formulate surgical plans based on the preoperative 3D geometries of the face and skull. Specifically, the surgical plan is defined by geometrically editing the skull geometry to achieve the desired postoperative outcome. A key step in this process is predicting the postoperative facial geometry corresponding to the modified skull, as it provides critical guidance for surgical decision-making and facilitates effective doctor-patient communication. We define the task of surgical planning prediction as follows: given a preoperative face \( \mathbf{F}_{\mathrm{before}} \), a preoperative skull \( \mathbf{S}_{\mathrm{before}} \), and a clinician-specified skull modification plan \( \mathbf{S}_{\mathrm{plan}} \), the goal is to predict the postoperative facial geometry \( \mathbf{F}_{\mathrm{after}}^{\mathrm{pred}} \). The following describes how our method achieves this.

\begin{figure*}[!htbp]
\centering
\includegraphics[width=0.95\textwidth]{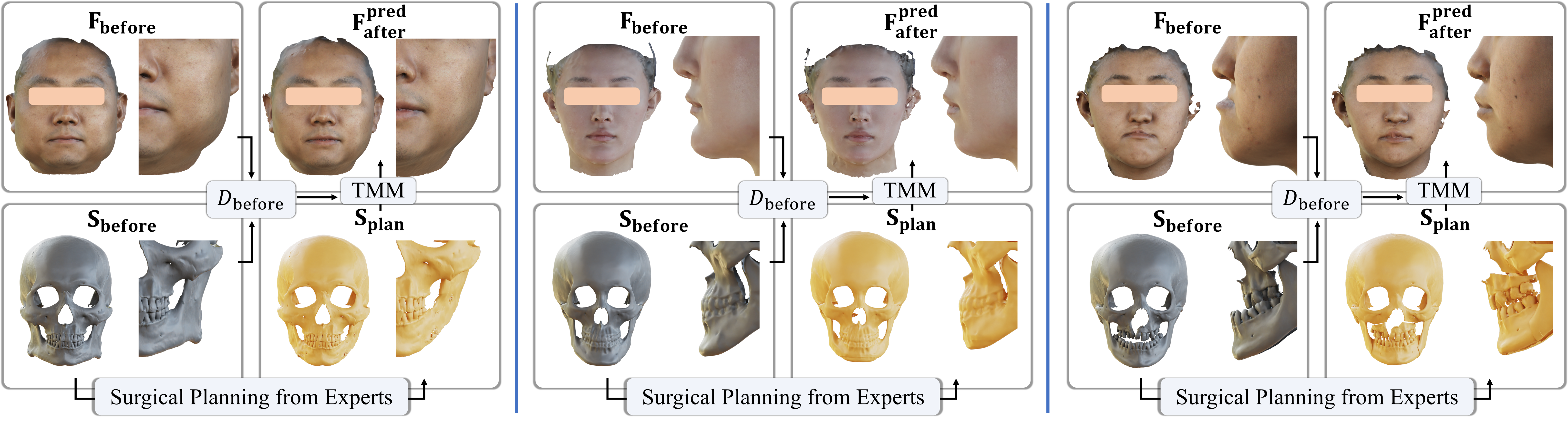}
\caption{Our Vision Tomography supports surgical planning prediction for aesthetic facial procedures, providing critical guidance for surgical decision-making and facilitating effective doctor-patient communication.
}
\label{Surgical_Planning_Prediction}
\end{figure*}

We first perform a topology registration of the preoperative face \( \mathbf{F}_{\mathrm{before}} \) and skull \( \mathbf{S}_{\mathrm{before}} \), consistent with Equations~\ref{face_Deformation} and~\ref{skull_Deformation}, which will not be repeated here, and no new notation is introduced. As shown in Figure~\ref{Surgical_Planning_Prediction}, our method estimates the tissue thickness \( {D}_{\mathrm{before}} \) between \( \mathbf{F}_{\mathrm{before}} \) and \( \mathbf{S}_{\mathrm{before}} \), the tissue thickness \( D_{\mathrm{before}} \) needs to be adjusted according to the changes in the surface normals from the preoperative skull \( \mathbf{S}_{\mathrm{before}} \) to the planned skull \( \mathbf{S}_{\mathrm{plan}} \), in order to fully exploit the geometric priors of the deformation. The adjustment is performed as follows: given the tissue thickness vector ${D}_{\mathrm{before}}$ defined with respect to the surface mesh $\mathbf{S}_{\mathrm{before}}$, we update it on the deformed mesh $\mathbf{S}_{\mathrm{plan}}$ by transferring the local displacement in a normal-tangent frame. Each vector in \( {D}_{\mathrm{before}} \) is decomposed into normal and tangential components on \( \mathbf{S}_{\mathrm{before}} \), and these components are preserved while reconstructing a new tissue thickness vector in the corresponding local frame of \( \mathbf{S}_{\mathrm{plan}} \), yielding the updated tissue thickness vector \( {D}_{\mathrm{plan}} \).

Then our method uses the updated \( {D}_{\mathrm{before}} \) (\( {D}_{\mathrm{plan}} \)) along with planned skull \( \mathbf{S}_{\mathrm{plan}} \) to predict the postoperative face \( \mathbf{F}_{\mathrm{after}}^{\mathrm{pred}} \). Specifically, similar to Equation~\ref{fit_TMM} in TMM, we first apply the updated \( {D}_{\mathrm{before}} \) (\( {D}_{\mathrm{plan}} \)) to the vertices of the skull modification plan
\( V_{{\mathrm{plan}}}\!\left[\mathcal{I}_{\mathrm{t}}\right] \in \mathbb{R}^{(N_{\mathrm{F}} \cdot N_q) \times 3} \) (\(V_{{\mathrm{plan}}}  \in \mathbf{S}_{\mathrm{plan}}\))
to estimate an initial prediction of postoperative face vertices
\( V'_{{\mathrm{pred}}} \in \mathbb{R}^{(N_{\mathrm{F}} \cdot N_q) \times 3} \). Subsequently, we refine these initial vertices using \( \boldsymbol{\Phi}_{\mathrm{fit}}(\cdot) \) to obtain the predicted postoperative face
\( {\mathbf{F}_{\mathrm{after}}^{\mathrm{pred}}}' \). Due to the regional limitations defined by the TMM and FSMM, \( {\mathbf{F}_{\mathrm{after}}^{\mathrm{pred}}}' \) only covers the face region. However, the original preoperative face scan \( \mathbf{F}_{\mathrm{before}} \) often includes additional boundary regions, such as parts of the neck. To ensure that the predicted postoperative head region consists with the preoperative scan, we use \( {\mathbf{F}_{\mathrm{after}}^{\mathrm{pred}}}' \) as control points and deform \( \mathbf{F}_{\mathrm{before}} \) using ARAP \cite{sorkine2007rigid}, resulting in the final surgical planning face prediction \( {\mathbf{F}_{\mathrm{after}}^{\mathrm{pred}}} \) that matches the original region.

The results of surgical planning prediction using VT are shown in Figure~\ref{Surgical_Planning_Prediction}, illustrating predicted postoperative facial geometry for patients undergoing facial slimming or orthognathic procedures, based on surgeon-specified modifications to the preoperative skull.

\begin{table}
\begin{center}
\caption{Ablation study for skull registration. Distance errors are measured in millimeters.}
\label{AblationStudytable}
\setlength{\tabcolsep}{0.6mm} 
\resizebox{0.9\columnwidth}{!}{
\begin{tabular}{cc|cc|cc}
\toprule[1pt]
\multicolumn{2}{c|}{$N_{q} = 1$ } & \multicolumn{2}{c|}{$N_{q} = 3$ } & \multicolumn{2}{c}{$N_{q} = 5$ }  \\ 
$N_{O}$  & RMSE $\downarrow$  & $N_{O}$ & RMSE $\downarrow$  & $N_{O}$ & RMSE $\downarrow$  \\ \midrule[0.5pt]
$N_\mathrm{F}/100$             & 0.600 & $N_\mathrm{F}/100$             & 0.599 & $N_\mathrm{F}/100$             & 0.598 \\
$N_\mathrm{F}/10$               & 0.596 & $N_\mathrm{F}/10$               & 0.594 & $N_\mathrm{F}/10$              & 0.592 \\
$N_\mathrm{F}$                    & 0.594 & $N_\mathrm{F}$                 &\textbf{ 0.588} & $N_\mathrm{F}$                   & 0.590 \\ \bottomrule[1pt]
\end{tabular}
}
\end{center}
\end{table}

\begin{figure}[t]
\centering
\includegraphics[width=0.8\columnwidth]{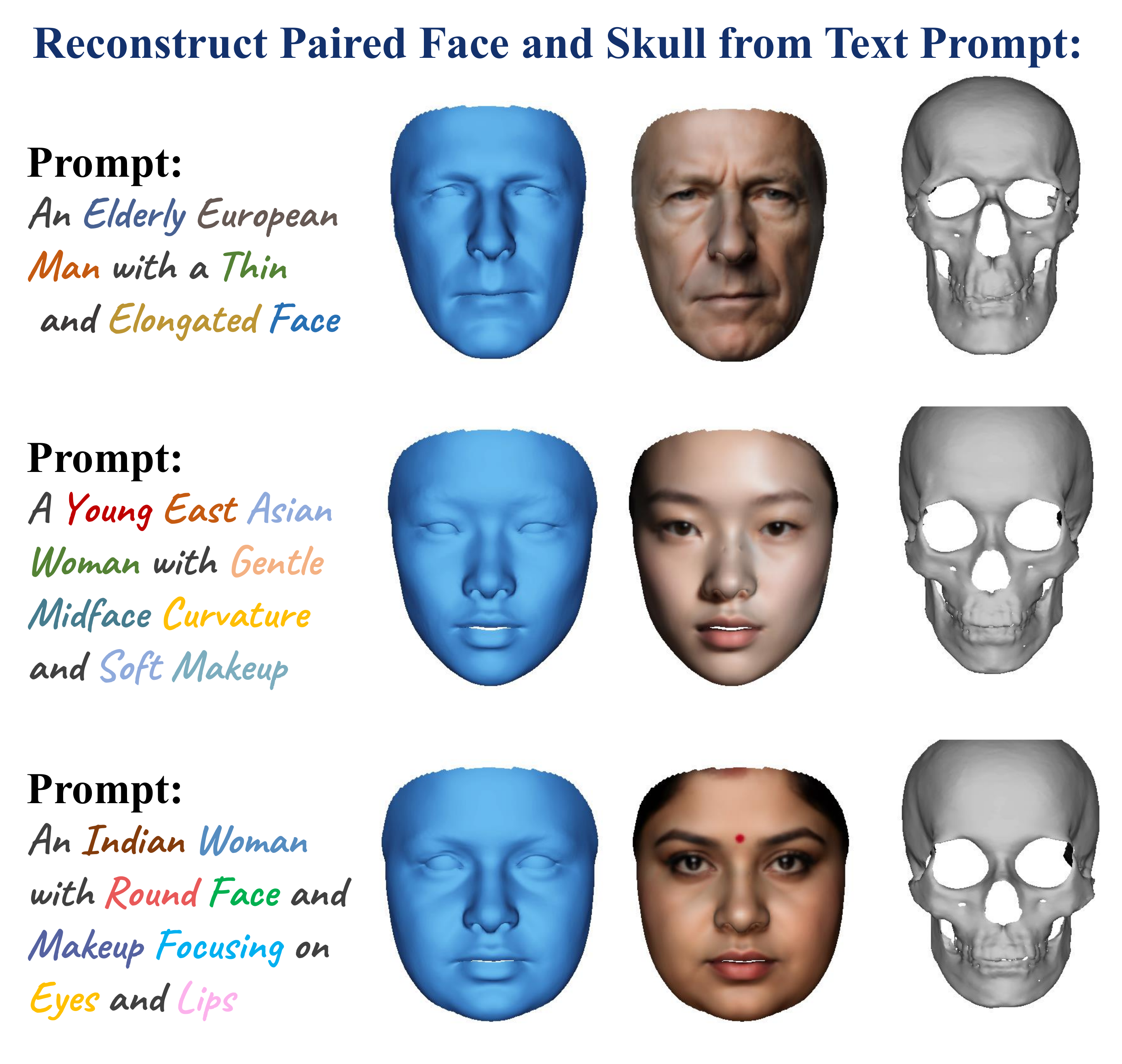}
\caption{Reconstruct paired face and skull from text prompt.}
\label{text3D}
\end{figure}

\begin{figure*}[!t]
\begin{center}
\includegraphics[width=0.95\linewidth]{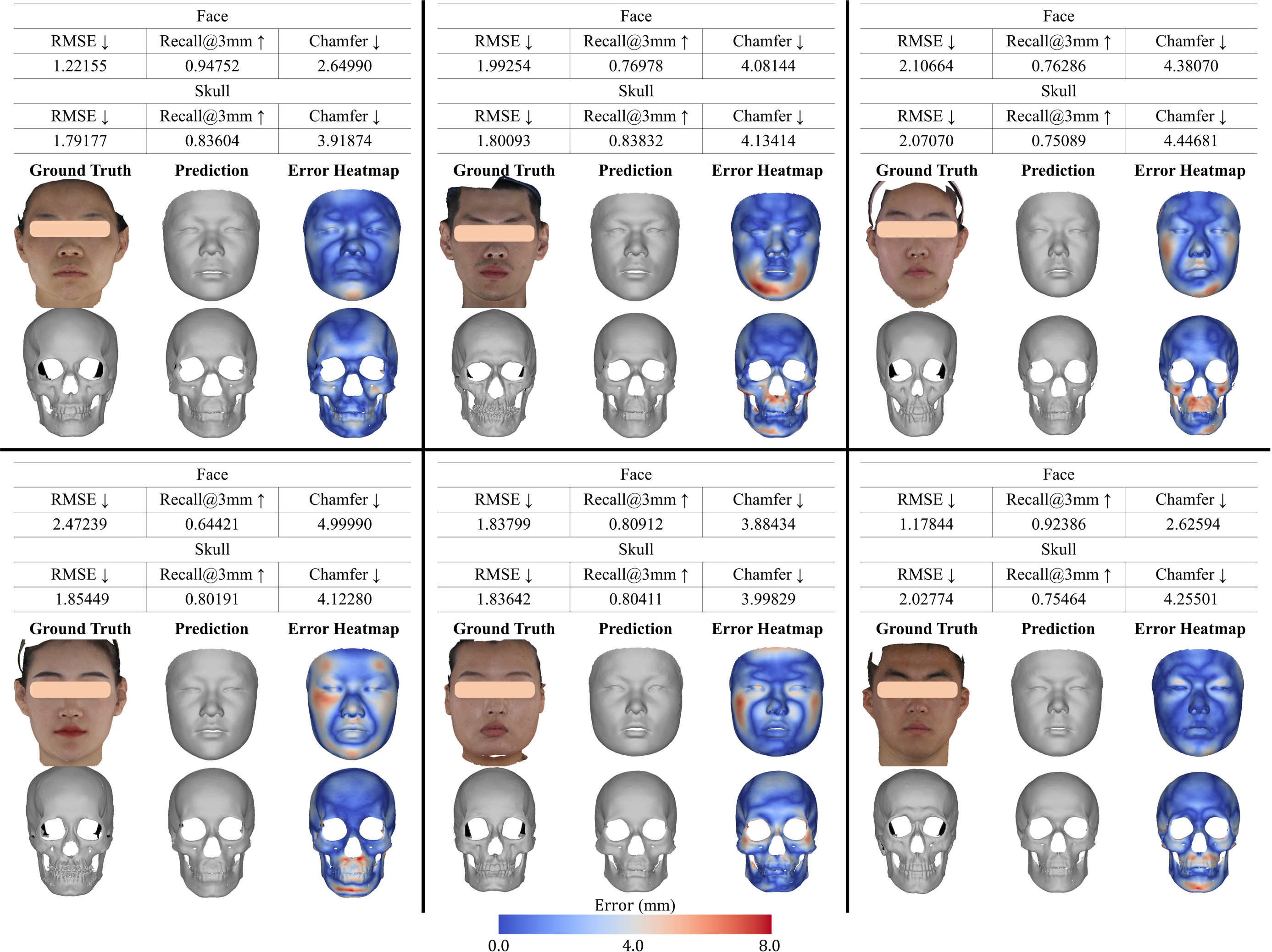}
\end{center}
\caption{Our method accurately reconstructs paired face and skull geometry from a single image. During testing, the top-left face scan of each example is rendered as image input to the Vision Tomography framework.
}
\label{skull_heatmap}
\end{figure*}

\section{Experiments}

\subsection{Experimental Settings}
\myparagraph{Data.} We use 161 identities (322 samples with flipping) to build the FSMM and TMM. For the remaining samples, we randomly select 15 (30 with flipping) for quantitative comparison. For the comparison of image-based reconstruction accuracy, we render the test samples from multiple viewpoints to generate input for the evaluated methods. To train \( \boldsymbol{\Phi}_\mathrm{recon}^I \), 2D in-the-wild face images from publicly available datasets \cite{liu2015faceattributes, sagonas2013300, CelebAMask-HQ} with pose augmentation \cite{zhu2017face} are used, excluding those with facial expressions to better align with FSMM. The training set for \( \boldsymbol{\Phi}_\mathrm{recon}^I \) contains approximately $150K$ face images. To train \( \boldsymbol{\Phi}_\mathrm{recon}^S \), we mainly follow \cite{wang2024s2td}, removing the fine-grained face reconstruction step to simplify the pipeline, similarly excluding images with facial expressions, and replacing the face model \cite{blanz1999morphable} used in \cite{wang2024s2td} with our FSMM.

\myparagraph{Implementation Details.} We implement Vision Tomography mainly based on PyTorch \cite{paszke2019pytorch}, Nvdiffrast \cite{Laine2020diffrast}, PyTorch3D \cite{ravi2020pytorch3d} and Kaolin \cite{KaolinLibrary}. The input image or sketch for \( \boldsymbol{\Phi}_\mathrm{recon}^I \) or \( \boldsymbol{\Phi}_\mathrm{recon}^S \) is resized into $224\cdot 224$. \(N_\mathrm{F} = 35709\). \(N_\mathrm{S} = 23208\).

\myparagraph{SOTA Methods for Comparison.} To evaluate the accuracy of paired reconstruction of the face and skull from images, we compare our method with state-of-the-art approaches \cite{SMIRK,DECA:Siggraph2021,wang20243d,deng2019accurate}. As these methods \cite{SMIRK,DECA:Siggraph2021,wang20243d,deng2019accurate} reconstruct only facial geometry, we extend their outputs using FSMM to estimate the skull, enabling a comprehensive evaluation of face-skull reconstruction. To evaluate the upper bound of BFSM in 3D face representation, we compare our model against state-of-the-art morphable models, including BFM \cite{blanz1999morphable}, and FLAME \cite{FLAME:SiggraphAsia2017}. FaceWarehouse \cite{cao2013facewarehouse} is used as the expression basis for BFM to enhance its representational capacity on certain slight deformity cases, enabling a fair comparison with BFSM.

\myparagraph{Metrics.} Accuracy is evaluated using Root Mean Square Error (RMSE), Chamfer Distance, and Recall. For Chamfer Distance, the GT is cropped to match the predicted region. Recall \cite{tatarchenko2019single, giebenhain2024mononphm} quantifies the percentage of GT points that have a nearby point in the reconstruction within a threshold.

\subsection{Visualization for Qualitative Analysis}

Figure~\ref{recon_res1} presents results of directly reconstructing anatomically aligned 3D face and skull geometry from images using VT. Figure~\ref{recon_res2} shows results of reconstructing paired 3D face and skull geometry from sketches of different styles, with the added ability to control the texture of the reconstructed 3D face based on descriptions. VT captures both the facial and underlying skull geometry in an identity-consistent manner and provides a portable, non-invasive, and radiation-free solution for obtaining craniofacial anatomical structures, which was not achievable with previous methods. As shown in Figure~\ref{text3D}, we further explore applications of text-driven 3D face and skull reconstruction. Given text descriptions \( T_X \) for a specific face, we leverage existing powerful text-to-image models \cite{flux1kreadev2025,flux-2-2025,labs2025flux1kontextflowmatching,flux2024} to first generate the corresponding 2D face image based on the text descriptions \( T_X \), and then use \( \boldsymbol{\Phi}_\mathrm{recon}^I \) to complete the reconstruction of an anatomically aligned 3D face-skull pair. These techniques and applications provide practical value for remote diagnostics, digital healthcare, forensic identification, missing-person investigations, and archaeological analysis by enabling non-invasive, portable, and identity-consistent reconstruction of craniofacial structures.

To demonstrate VT’s capability for joint editing of face and skull geometry, we apply it to surgical planning prediction. Given preoperative facial and skull geometry along with the surgeon-edited skull (representing the planned surgical modifications), our method predicts the postoperative facial geometry. This technology provides critical guidance for surgical decision-making and facilitates effective doctor–patient communication. Figure~\ref{Surgical_Planning_Prediction} presents examples of surgical planning prediction results across various clinical aesthetic cases. These cosmetic surgery plans typically involve geometric edits to the zygomatic bones, mandible, and the midline region of the upper lip (philtrum), aiming to achieve facial slimming or correct minor craniofacial deformities. The results demonstrate that our method can accurately predict postoperative facial geometry while preserving identity consistency and reflecting the intended surgical modifications.

\subsection{Quantitative Comparison}
\label{subsec:QuantitativeComparison}
Table~\ref{Inference-based} presents the quantitative comparison of image-based reconstruction accuracy, where our method achieves the best overall performance. Note that skull reconstructions for compared methods are obtained by applying our FSMM to their respective face reconstruction outputs. Figure~\ref{skull_heatmap} presents the error heatmap results of our VT for image-based 3D face and skull reconstruction. 

The 3D Bidirectional Face-Skull Morphable Model (BFSM) lies at the core of VT. To evaluate its upper bound in representing diverse facial geometries, we deform various morphable models \cite{blanz1999morphable,FLAME:SiggraphAsia2017,cao2013facewarehouse} to fit a set of 3D face samples and measure face reconstruction errors against the ground-truth meshes. Table~\ref{optimize-based} reports these upper bounds, highlighting the advantages of our method. Each model is optimized to minimize reconstruction errors, reflecting the best representation it can achieve. Consequently, the errors in Table~\ref{optimize-based} are lower than those in Table~\ref{Inference-based}.

\subsection{Ablation Study}
\label{AblationStudy}

During the dense ray matching topology registration of the face, skull, and tissue thickness, \textit{i.e.}, when applying Equations~\ref{dense_skull_ray} and~\ref{skull_Deformation}, the ray origins and ray endpoints determine thousands of semantically consistent control point pairs for non-rigid deformation. We conducted an ablation study to evaluate the impact of the number of ray origins and ray endpoints on the accuracy of skull registration. Table~\ref{AblationStudytable} shows that increasing the number of ray origins (\(N_O\)) may improve registration accuracy by using denser control point correspondences, while setting the number of ray endpoints to \(N_q = 3\) yields the best overall performance.

\section{Conclusions}

This paper proposes Vision Tomography, a portable and non-invasive framework capable of 3D face-skull reconstruction from various visual data, and also supporting joint editing of face and skull geometry for surgical planning prediction. Its main contributions include the introduction of a large face-skull dataset and a novel dense ray matching registration method that ensures topological consistency across different cases. The core of Vision Tomography is the 3D Bidirectional Face-Skull Morphable Model (BFSM), which enables accurate bidirectional inference between face and skull geometry and models tissue variability for diverse reconstructions. Experiments demonstrate that our framework outperforms existing state-of-the-art methods and shows promise for a wide range of applications.

\bibliographystyle{IEEEtran}  
\bibliography{skull}

@inproceedings{blanz1999morphable,
  title={A morphable model for the synthesis of 3D faces},
  author={Blanz, Volker and Vetter, Thomas},
  booktitle={Proceedings of the 26th annual conference on Computer graphics and interactive techniques},
  pages={187--194},
  year={1999}
}

@article{blanz2003face,
  title={Face recognition based on fitting a 3D morphable model},
  author={Blanz, Volker and Vetter, Thomas},
  journal={IEEE Transactions on pattern analysis and machine intelligence},
  volume={25},
  number={9},
  pages={1063--1074},
  year={2003},
  publisher={IEEE}
}

@article{zhu2017face,
  title={Face alignment in full pose range: A 3d total solution},
  author={Zhu, Xiangyu and Liu, Xiaoming and Lei, Zhen and Li, Stan Z},
  journal={IEEE transactions on pattern analysis and machine intelligence},
  volume={41},
  number={1},
  pages={78--92},
  year={2017},
  publisher={IEEE}
}

@article{FLAME:SiggraphAsia2017, 
  title = {Learning a model of facial shape and expression from {4D} scans}, 
  author = {Li, Tianye and Bolkart, Timo and Black, Michael. J. and Li, Hao and Romero, Javier}, 
  journal = {ACM Transactions on Graphics, (Proc. SIGGRAPH Asia)}, 
  volume = {36}, 
  number = {6}, 
  year = {2017}, 
  pages = {194:1--194:17},
  url = {https://doi.org/10.1145/3130800.3130813} 
}

@InProceedings{yang2020facescape,
  author = {Yang, Haotian and Zhu, Hao and Wang, Yanru and Huang, Mingkai and Shen, Qiu and Yang, Ruigang and Cao, Xun},
  title = {FaceScape: A Large-Scale High Quality 3D Face Dataset and Detailed Riggable 3D Face Prediction},
  booktitle = {IEEE/CVF Conference on Computer Vision and Pattern Recognition (CVPR)},
  month = {June},
  year = {2020},
  page = {601--610}}

@inproceedings{wang2022faceverse,
  title={Faceverse: a fine-grained and detail-controllable 3d face morphable model from a hybrid dataset},
  author={Wang, Lizhen and Chen, Zhiyuan and Yu, Tao and Ma, Chenguang and Li, Liang and Liu, Yebin},
  booktitle={Proceedings of the IEEE/CVF Conference on Computer Vision and Pattern Recognition},
  pages={20333--20342},
  year={2022}
}

@inproceedings{DECA:Siggraph2021,
  title={Learning an Animatable Detailed {3D} Face Model from In-The-Wild Images},
  author={Feng, Yao and Feng, Haiwen and Black, Michael J. and Bolkart, Timo},
  journal = {ACM Transactions on Graphics, (Proc. SIGGRAPH)}, 
  volume = {40}, 
  number = {8}, 
  year = {2021}, 
  url = {https://doi.org/10.1145/3450626.3459936} 
}

@inproceedings{CelebAMask-HQ,
  title={MaskGAN: Towards Diverse and Interactive Facial Image Manipulation},
  author={Lee, Cheng-Han and Liu, Ziwei and Wu, Lingyun and Luo, Ping},
  booktitle={IEEE Conference on Computer Vision and Pattern Recognition (CVPR)},
  year={2020}
}

@inproceedings{liu2015faceattributes,
  title = {Deep Learning Face Attributes in the Wild},
  author = {Liu, Ziwei and Luo, Ping and Wang, Xiaogang and Tang, Xiaoou},
  booktitle = {Proceedings of International Conference on Computer Vision (ICCV)},
  month = {December},
  year = {2015} 
}

@misc{nicp_github,
  author = {},
  title = {GPU Accelerated Non-rigid ICP for surface registration},
  year = {2021},
  publisher = {GitHub},
  journal = {GitHub repository},
  howpublished = {\url{https://github.com/wuhaozhe/pytorch-nicp}}
}

@article{ravi2020pytorch3d,
    author = {Nikhila Ravi and Jeremy Reizenstein and David Novotny and Taylor Gordon
                  and Wan-Yen Lo and Justin Johnson and Georgia Gkioxari},
    title = {Accelerating 3D Deep Learning with PyTorch3D},
    journal = {arXiv:2007.08501},
    year = {2020},
}

@inproceedings{sagonas2013300,
  title={300 faces in-the-wild challenge: The first facial landmark localization challenge},
  author={Sagonas, Christos and Tzimiropoulos, Georgios and Zafeiriou, Stefanos and Pantic, Maja},
  booktitle={Proceedings of the IEEE international conference on computer vision workshops},
  pages={397--403},
  year={2013}
}

@article{paszke2019pytorch,
  title={Pytorch: An imperative style, high-performance deep learning library},
  author={Paszke, Adam and Gross, Sam and Massa, Francisco and Lerer, Adam and Bradbury, James and Chanan, Gregory and Killeen, Trevor and Lin, Zeming and Gimelshein, Natalia and Antiga, Luca and others},
  journal={Advances in neural information processing systems},
  volume={32},
  year={2019}
}

@inproceedings{deng2019accurate,
  title={Accurate 3d face reconstruction with weakly-supervised learning: From single image to image set},
  author={Deng, Yu and Yang, Jiaolong and Xu, Sicheng and Chen, Dong and Jia, Yunde and Tong, Xin},
  booktitle={Proceedings of the IEEE/CVF conference on computer vision and pattern recognition workshops},
  pages={0--0},
  year={2019}
}

@article{dai2020statisticalLYHM,
  title={Statistical modeling of craniofacial shape and texture},
  author={Dai, Hang and Pears, Nick and Smith, William and Duncan, Christian},
  journal={International Journal of Computer Vision},
  volume={128},
  pages={547--571},
  year={2020},
  publisher={Springer}
}

@misc{KaolinLibrary,
      author = {Fuji Tsang, Clement and Shugrina, Maria and Lafleche, Jean Francois and Takikawa, Towaki and Wang, Jiehan and Loop, Charles and Chen, Wenzheng and Jatavallabhula, Krishna Murthy and Smith, Edward and Rozantsev, Artem and Perel, Or and Shen, Tianchang and Gao, Jun and Fidler, Sanja and State, Gavriel and Gorski, Jason and Xiang, Tommy and Li, Jianing and Li, Michael and Lebaredian, Rev},
      title = {Kaolin: A Pytorch Library for Accelerating 3D Deep Learning Research},
      year = {2022},
      howpublished={\url{https://github.com/NVIDIAGameWorks/kaolin}}
}

@inproceedings{amberg2007optimal,
  title={Optimal step nonrigid ICP algorithms for surface registration},
  author={Amberg, Brian and Romdhani, Sami and Vetter, Thomas},
  booktitle={2007 IEEE conference on computer vision and pattern recognition},
  pages={1--8},
  year={2007},
  organization={IEEE}
}

@inproceedings{li2020learning,
title={Learning Formation of Physically-Based Face Attributes},
author={Ruilong Li and Karl Bladin and Yajie Zhao and Chinmay Chinara and Owen Ingraham and Pengda Xiang and Xinglei Ren and Pratusha Prasad and Bipin Kishore and Jun Xing and Hao Li},
year={2020},
eprint={2004.03458},
archivePrefix={arXiv},
primaryClass={cs.CV}
}

@article{bradski2000opencv,
  title={The openCV library.},
  author={Bradski, Gary},
  journal={Dr. Dobb's Journal: Software Tools for the Professional Programmer},
  volume={25},
  number={11},
  pages={120--123},
  year={2000},
  publisher={Miller Freeman Inc.}
}

@article{simo2018mastering,
  title={Mastering sketching: adversarial augmentation for structured prediction},
  author={Simo-Serra, Edgar and Iizuka, Satoshi and Ishikawa, Hiroshi},
  journal={ACM Transactions on Graphics (TOG)},
  volume={37},
  number={1},
  pages={1--13},
  year={2018},
  publisher={ACM New York, NY, USA}
}

@article{simo2016learning,
  title={Learning to simplify: fully convolutional networks for rough sketch cleanup},
  author={Simo-Serra, Edgar and Iizuka, Satoshi and Sasaki, Kazuma and Ishikawa, Hiroshi},
  journal={ACM Transactions on Graphics (TOG)},
  volume={35},
  number={4},
  pages={1--11},
  year={2016},
  publisher={ACM New York, NY, USA}
}

@misc{zhang2023adding,
  title={Adding Conditional Control to Text-to-Image Diffusion Models}, 
  author={Lvmin Zhang and Anyi Rao and Maneesh Agrawala},
  booktitle={IEEE International Conference on Computer Vision (ICCV)},
  year={2023},
}

@misc{flux1kreadev2025,
    author={Sangwu Lee and Titus Ebbecke and Erwann Millon and Will Beddow and Le Zhuo and Iker García-Ferrero and Liam Esparraguera and Mihai Petrescu and Gian Saß and Gabriel Menezes and Victor Perez},
    title={FLUX.1 Krea [dev]},
    year={2025},
    howpublished={\url{https://github.com/krea-ai/flux-krea}},
}

@misc{flux-2-2025,
    author={Black Forest Labs},
    title={{FLUX.2: Frontier Visual Intelligence}},
    year={2025},
    howpublished={\url{https://bfl.ai/blog/flux-2}},
}

@misc{labs2025flux1kontextflowmatching,
      title={FLUX.1 Kontext: Flow Matching for In-Context Image Generation and Editing in Latent Space},
      author={Black Forest Labs and Stephen Batifol and Andreas Blattmann and Frederic Boesel and Saksham Consul and Cyril Diagne and Tim Dockhorn and Jack English and Zion English and Patrick Esser and Sumith Kulal and Kyle Lacey and Yam Levi and Cheng Li and Dominik Lorenz and Jonas Müller and Dustin Podell and Robin Rombach and Harry Saini and Axel Sauer and Luke Smith},
      year={2025},
      eprint={2506.15742},
      archivePrefix={arXiv},
      primaryClass={cs.GR},
      url={https://arxiv.org/abs/2506.15742},
}

@misc{flux2024,
    author={Black Forest Labs},
    title={FLUX},
    year={2024},
    howpublished={\url{https://github.com/black-forest-labs/flux}},
}

@inproceedings{wood2021fake,
  title={Fake it till you make it: face analysis in the wild using synthetic data alone},
  author={Wood, Erroll and Baltru{\v{s}}aitis, Tadas and Hewitt, Charlie and Dziadzio, Sebastian and Cashman, Thomas J and Shotton, Jamie},
  booktitle={Proceedings of the IEEE/CVF international conference on computer vision},
  pages={3681--3691},
  year={2021}
}

@inproceedings{sorkine2007rigid,
  title={As-rigid-as-possible surface modeling},
  author={Sorkine, Olga and Alexa, Marc},
  booktitle={Symposium on Geometry processing},
  volume={4},
  pages={109--116},
  year={2007}
}

@inproceedings{ramamoorthi2001efficient,
  title={An efficient representation for irradiance environment maps},
  author={Ramamoorthi, Ravi and Hanrahan, Pat},
  booktitle={Proceedings of the 28th annual conference on Computer graphics and interactive techniques},
  pages={497--500},
  year={2001}
}

@inproceedings{mildenhall2020nerf,
  title={NeRF: Representing Scenes as Neural Radiance Fields for View Synthesis},
  author={Ben Mildenhall and Pratul P. Srinivasan and Matthew Tancik and Jonathan T. Barron and Ravi Ramamoorthi and Ren Ng},
  year={2020},
  booktitle={ECCV},
}

@inproceedings{giebenhain2023nphm,
 author={Simon Giebenhain and Tobias Kirschstein and Markos Georgopoulos and  Martin R{\"{u}}nz and Lourdes Agapito and Matthias Nie{\ss}ner},
 title={Learning Neural Parametric Head Models},
 booktitle = {Proc. IEEE Conf. on Computer Vision and Pattern Recognition (CVPR)},
 year = {2023}}

@inproceedings{wang2024s2td,
  title={S2TD-Face: Reconstruct a Detailed 3D Face with Controllable Texture from a Single Sketch},
  author={Wang, Zidu and Zhu, Xiangyu and Yu, Jiang and Zhang, Tianshuo and Lei, Zhen},
  booktitle={Proceedings of the 32nd ACM International Conference on Multimedia},
  pages={6453--6462},
  year={2024}
}

@inproceedings{wang20243d,
  title={3D Face Reconstruction with the Geometric Guidance of Facial Part Segmentation},
  author={Wang, Zidu and Zhu, Xiangyu and Zhang, Tianshuo and Wang, Baiqin and Lei, Zhen},
  booktitle={Proceedings of the IEEE/CVF Conference on Computer Vision and Pattern Recognition},
  pages={1672--1682},
  year={2024}
}

@article{Sirovich:87,
author = {L. Sirovich and M. Kirby},
journal = {J. Opt. Soc. Am. A},
number = {3},
pages = {519--524},
publisher = {Optica Publishing Group},
title = {Low-dimensional procedure for the characterization of human faces},
volume = {4},
month = {Mar},
year = {1987},
url = {https://opg.optica.org/josaa/abstract.cfm?URI=josaa-4-3-519},
doi = {10.1364/JOSAA.4.000519},
}

@article{jolliffe2016principal,
  title={Principal component analysis: a review and recent developments},
  author={Jolliffe, Ian T and Cadima, Jorge},
  journal={Philosophical transactions of the royal society A: Mathematical, Physical and Engineering Sciences},
  volume={374},
  number={2065},
  pages={20150202},
  year={2016},
  publisher={the Royal Society publishing}
}

@inproceedings{SMIRK,
  title={3D facial expressions through analysis-by-neural-synthesis},
  author={Retsinas, George and Filntisis, Panagiotis P and Danecek, Radek and Abrevaya, Victoria F and Roussos, Anastasios and Bolkart, Timo and Maragos, Petros},
  booktitle={Proceedings of the IEEE/CVF Conference on Computer Vision and Pattern Recognition},
  pages={2490--2501},
  year={2024}
}

@article{Laine2020diffrast,
  title   = {Modular Primitives for High-Performance Differentiable Rendering},
  author  = {Samuli Laine and Janne Hellsten and Tero Karras and Yeongho Seol and Jaakko Lehtinen and Timo Aila},
  journal = {ACM Transactions on Graphics},
  year    = {2020},
  volume  = {39},
  number  = {6}
}

@inProceedings{giebenhain2024mononphm,
 author={Simon Giebenhain and Tobias Kirschstein and Markos Georgopoulos and  Martin R{\"{u}}nz and Lourdes Agapito and Matthias Nie{\ss}ner},
 title={MonoNPHM: Dynamic Head Reconstruction from Monoculuar Videos},
 booktitle = {Proc. IEEE Conf. on Computer Vision and Pattern Recognition (CVPR)},
 year = {2024}
 }

@inproceedings{tatarchenko2019single,
  title={What do single-view 3d reconstruction networks learn?},
  author={Tatarchenko, Maxim and Richter, Stephan R and Ranftl, Ren{\'e} and Li, Zhuwen and Koltun, Vladlen and Brox, Thomas},
  booktitle={Proceedings of the IEEE/CVF conference on computer vision and pattern recognition},
  pages={3405--3414},
  year={2019}
}

@ARTICLE{10127617,
  author={Kao, Yueying and Pan, Bowen and Xu, Miao and Lyu, Jiangjing and Zhu, Xiangyu and Chang, Yuanzhang and Li, Xiaobo and Lei, Zhen},
  journal={IEEE Transactions on Image Processing}, 
  title={Toward 3D Face Reconstruction in Perspective Projection: Estimating 6DoF Face Pose From Monocular Image}, 
  year={2023},
  volume={32},
  number={},
  pages={3080-3091},
  doi={10.1109/TIP.2023.3275535}}

@ARTICLE{10506678,
  author={Xu, Miao and Zhu, Xiangyu and Kao, Yueying and Chen, Zhiwen and Lyu, Jiangjing and Lei, Zhen},
  journal={IEEE Transactions on Multimedia}, 
  title={Multi-Level Pixel-Wise Correspondence Learning for 6DoF Face Pose Estimation}, 
  year={2024},
  volume={26},
  number={},
  pages={9423-9435},
  doi={10.1109/TMM.2024.3391888}}

@article{reddi2019convergence,
  title={On the convergence of adam and beyond},
  author={Reddi, Sashank J and Kale, Satyen and Kumar, Sanjiv},
  journal={arXiv preprint arXiv:1904.09237},
  year={2019}
}

@article{zoss2019accurate,
  author       = {Zoss, G. and Beeler, T. and Gross, M. and others},
  title        = {Accurate Markerless Jaw Tracking for Facial Performance Capture},
  journal      = {ACM Transactions on Graphics},
  year         = {2019},
  volume       = {38},
  number       = {4},
  pages        = {1--8},
  doi          = {10.1145/3306346.3322968}
}

@inproceedings{yang2023implicit,
  author       = {Yang, L. and Zoss, G. and Chandran, P. and others},
  title        = {An Implicit Physical Face Model Driven by Expression and Style},
  booktitle    = {SIGGRAPH Asia 2023 Conference Papers},
  year         = {2023},
  pages        = {1--12},
  location     = {Sydney, Australia},
  doi          = {10.1145/3610548.3618223}
}

@misc{wang2025srmhair,
      title={SRM-Hair: Single Image Head Mesh Reconstruction via 3D Morphable Hair}, 
      author={Zidu Wang and Jiankuo Zhao and Miao Xu and Xiangyu Zhu and Zhen Lei},
      year={2025},
      eprint={2503.06154},
      archivePrefix={arXiv},
      primaryClass={cs.CV},
      url={https://arxiv.org/abs/2503.06154}, 
}

@inproceedings{keller2024hit,
  title={HIT: Estimating internal human implicit tissues from the body surface},
  author={Keller, Marilyn and Arora, Vaibhav and Dakri, Abdelmouttaleb and Chandhok, Shivam and Machann, J{\"u}rgen and Fritsche, Andreas and Black, Michael J and Pujades, Sergi},
  booktitle={Proceedings of the IEEE/CVF Conference on Computer Vision and Pattern Recognition},
  pages={3480--3490},
  year={2024}
}

@article{keller2023skin,
  title={From skin to skeleton: Towards biomechanically accurate 3d digital humans},
  author={Keller, Marilyn and Werling, Keenon and Shin, Soyong and Delp, Scott and Pujades, Sergi and Liu, C Karen and Black, Michael J},
  journal={ACM Transactions on Graphics (TOG)},
  volume={42},
  number={6},
  pages={1--12},
  year={2023},
  publisher={ACM New York, NY, USA}
}

@inproceedings{keller2022osso,
  title={OSSO: Obtaining skeletal shape from outside},
  author={Keller, Marilyn and Zuffi, Silvia and Black, Michael J and Pujades, Sergi},
  booktitle={Proceedings of the IEEE/CVF conference on computer vision and pattern recognition},
  pages={20492--20501},
  year={2022}
}

@inproceedings{luthi2009hierarchical,
  title={A hierarchical, multi-resolution approach for model-based skull-segmentation in mri volumes},
  author={L{\"u}thi, Marcel and Lerch, Anita and Albrecht, Thomas and Krol, Zdzislaw and Vetter, Thomas},
  booktitle={Conference Proceedings D},
  volume={3},
  year={2009}
}

@inproceedings{paysan2009face,
  title={Face reconstruction from skull shapes and physical attributes},
  author={Paysan, Pascal and L{\"u}thi, Marcel and Albrecht, Thomas and Lerch, Anita and Amberg, Brian and Santini, Francesco and Vetter, Thomas},
  booktitle={Pattern Recognition: 31st DAGM Symposium, Jena, Germany, September 9-11, 2009. Proceedings 31},
  pages={232--241},
  year={2009},
  organization={Springer}
}

@inproceedings{madsen2018probabilistic,
  title={Probabilistic joint face-skull modelling for facial reconstruction},
  author={Madsen, Dennis and L{\"u}thi, Marcel and Schneider, Andreas and Vetter, Thomas},
  booktitle={Proceedings of the IEEE conference on computer vision and pattern recognition},
  pages={5295--5303},
  year={2018}
}

@inproceedings{achenbach2018multilinear,
  title={A Multilinear Model for Bidirectional Craniofacial Reconstruction.},
  author={Achenbach, Jascha and Brylka, Robert and Gietzen, Thomas and Zum Hebel, Katja and Sch{\"o}mer, Elmar and Schulze, Ralf and Botsch, Mario and Schwanecke, Ulrich},
  booktitle={VCBM@ MICCAI},
  pages={67--76},
  year={2018}
}

@ARTICLE{10791299SkulltoFace,
  author={Liang, Yongqing and Zhang, Congyi and Zhao, Junli and Wang, Wenping and Li, Xin},
  journal={IEEE Transactions on Visualization and Computer Graphics}, 
  title={Skull-to-Face: Anatomy-Guided 3D Facial Reconstruction and Editing}, 
  year={2024},
  volume={},
  number={},
  pages={1-13},
  doi={10.1109/TVCG.2024.3515093}
      }

@article{yang2024learning,
  title={Learning a Generalized Physical Face Model From Data},
  author={Yang, Lingchen and Zoss, Gaspard and Chandran, Prashanth and Gross, Markus and Solenthaler, Barbara and Sifakis, Eftychios and Bradley, Derek},
  journal={arXiv preprint arXiv:2402.19477},
  year={2024}
}

@article{yang2022implicit,
  title={Implicit neural representation for physics-driven actuated soft bodies},
  author={Yang, Lingchen and Kim, Byungsoo and Zoss, Gaspard and G{\"o}zc{\"u}, Baran and Gross, Markus and Solenthaler, Barbara},
  journal={ACM Transactions on Graphics (TOG)},
  volume={41},
  number={4},
  pages={1--10},
  year={2022},
  publisher={ACM New York, NY, USA}
}

@inproceedings{chandran2024anatomically,
  title={Anatomically Constrained Implicit Face Models},
  author={Chandran, Prashanth and Zoss, Gaspard},
  booktitle={Proceedings of the IEEE/CVF Conference on Computer Vision and Pattern Recognition},
  pages={2220--2229},
  year={2024}
}

@article{wang2021neus,
  title={Neus: Learning neural implicit surfaces by volume rendering for multi-view reconstruction},
  author={Wang, Peng and Liu, Lingjie and Liu, Yuan and Theobalt, Christian and Komura, Taku and Wang, Wenping},
  journal={arXiv preprint arXiv:2106.10689},
  year={2021}
}

@article{porto2021alpaca,
  title={ALPACA: A fast and accurate computer vision approach for automated landmarking of three-dimensional biological structures},
  author={Porto, Arthur and Rolfe, Sara and Maga, A Murat},
  journal={Methods in Ecology and Evolution},
  volume={12},
  number={11},
  pages={2129--2144},
  year={2021},
  publisher={Wiley Online Library}
}

@inproceedings{potamias2024shapefusion,
  title={Shapefusion: A 3d diffusion model for localized shape editing},
  author={Potamias, Rolandos Alexandros and Tarasiou, Michail and Ploumpis, Stylianos and Zafeiriou, Stefanos},
  booktitle={European Conference on Computer Vision},
  pages={72--89},
  year={2024},
  organization={Springer}
}

@misc{tarasiou2024locally,
      title={Locally Adaptive Neural 3D Morphable Models}, 
      author={Michail Tarasiou and Rolandos Alexandros Potamias and Eimear O'Sullivan and Stylianos Ploumpis and Stefanos Zafeiriou},
      year={2024},
      eprint={2401.02937},
      archivePrefix={arXiv},
      primaryClass={cs.CV}
}

@inproceedings{foti20223d,
  title={3d shape variational autoencoder latent disentanglement via mini-batch feature swapping for bodies and faces},
  author={Foti, Simone and Koo, Bongjin and Stoyanov, Danail and Clarkson, Matthew J},
  booktitle={Proceedings of the IEEE/CVF conference on computer vision and pattern recognition},
  pages={18730--18739},
  year={2022}
}

@article{cao2013facewarehouse,
  title={Facewarehouse: A 3d facial expression database for visual computing},
  author={Cao, Chen and Weng, Yanlin and Zhou, Shun and Tong, Yiying and Zhou, Kun},
  journal={IEEE Transactions on Visualization and Computer Graphics},
  volume={20},
  number={3},
  pages={413--425},
  year={2013},
  publisher={IEEE}
}

@article{SCULPTOR,
author = {Qiu, Zesong and Li, Yuwei and He, Dongming and Zhang, Qixuan and Zhang, Longwen and Zhang, Yinghao and Wang, Jingya and Xu, Lan and Wang, Xudong and Zhang, Yuyao and Yu, Jingyi},
title = {SCULPTOR: Skeleton-Consistent Face Creation Using a Learned Parametric Generator},
year = {2022},
issue_date = {December 2022},
publisher = {Association for Computing Machinery},
address = {New York, NY, USA},
volume = {41},
number = {6},
issn = {0730-0301},
journal = {ACM Trans. Graph.},
month = nov,
articleno = {213},
numpages = {17}
}

@article{Ichim_Phace,
author = {Ichim, Alexandru-Eugen and Kadle\v{c}ek, Petr and Kavan, Ladislav and Pauly, Mark},
title = {Phace: physics-based face modeling and animation},
year = {2017},
issue_date = {August 2017},
publisher = {Association for Computing Machinery},
address = {New York, NY, USA},
volume = {36},
number = {4},
issn = {0730-0301},
url = {https://doi.org/10.1145/3072959.3073664},
doi = {10.1145/3072959.3073664},
journal = {ACM Trans. Graph.},
month = jul,
articleno = {153},
numpages = {14}
}

@inproceedings{xu2024teethdreamer,
  title={TeethDreamer: 3D teeth reconstruction from five intra-oral photographs},
  author={Xu, Chenfan and Liu, Zhentao and Liu, Yuan and Dou, Yulong and Wu, Jiamin and Wang, Jiepeng and Wang, Minjiao and Shen, Dinggang and Cui, Zhiming},
  booktitle={International Conference on Medical Image Computing and Computer-Assisted Intervention},
  pages={712--721},
  year={2024},
  organization={Springer}
}

@article{zhang2022implicit,
  title={An implicit parametric morphable dental model},
  author={Zhang, Congyi and Elgharib, Mohamed and Fox, Gereon and Gu, Min and Theobalt, Christian and Wang, Wenping},
  journal={ACM Transactions on Graphics (TOG)},
  volume={41},
  number={6},
  pages={1--13},
  year={2022},
  publisher={ACM New York, NY, USA}
}

@article{bao2024deep,
  title={Deep learning--based facial and skeletal transformations for surgical planning},
  author={Bao, J and Zhang, X and Xiang, S and Liu, H and Cheng, M and Yang, Y and Huang, X and Xiang, W and Cui, W and Lai, HC and others},
  journal={Journal of Dental Research},
  volume={103},
  number={8},
  pages={809--819},
  year={2024},
  publisher={SAGE Publications Sage CA: Los Angeles, CA}
}

@article{zhang2025tcfnet,
  title={TCFNet: Bidirectional face-bone transformation via a Transformer-based coarse-to-fine point movement network},
  author={Zhang, Runshi and Jie, Bimeng and He, Yang and Wang, Junchen},
  journal={Medical Image Analysis},
  volume={105},
  pages={103653},
  year={2025},
  publisher={Elsevier}
}

@article{ma2023bidirectional,
  title={Bidirectional prediction of facial and bony shapes for orthognathic surgical planning},
  author={Ma, Lei and Lian, Chunfeng and Kim, Daeseung and Xiao, Deqiang and Wei, Dongming and Liu, Qin and Kuang, Tianshu and Ghanbari, Maryam and Li, Guoshi and Gateno, Jaime and others},
  journal={Medical image analysis},
  volume={83},
  pages={102644},
  year={2023},
  publisher={Elsevier}
}

@article{vaswani2017attention,
  title={Attention is all you need},
  author={Vaswani, Ashish and Shazeer, Noam and Parmar, Niki and Uszkoreit, Jakob and Jones, Llion and Gomez, Aidan N and Kaiser, {\L}ukasz and Polosukhin, Illia},
  journal={Advances in neural information processing systems},
  volume={30},
  year={2017}
}

@inproceedings{zhao2023fast,
  title={FAST-Net: A coarse-to-fine pyramid network for face-skull transformation},
  author={Zhao, Lei and Ma, Lei and Cui, Zhiming and Zheng, Jie and Xue, Zhong and Shi, Feng and Shen, Dinggang},
  booktitle={International Workshop on Machine Learning in Medical Imaging},
  pages={104--113},
  year={2023},
  organization={Springer}
}

@article{kim2025predicting,
  title={Predicting orthognathic surgery results as postoperative lateral cephalograms using graph neural networks and diffusion models},
  author={Kim, In-Hwan and Jeong, Jiheon and Kim, Jun-Sik and Lim, Jisup and Cho, Jin-Hyoung and Hong, Mihee and Kang, Kyung-Hwa and Kim, Minji and Kim, Su-Jung and Kim, Yoon-Ji and others},
  journal={Nature Communications},
  volume={16},
  number={1},
  pages={2586},
  year={2025},
  publisher={Nature Publishing Group UK London}
}

@article{du2024postoperative,
  title={Postoperative facial prediction for mandibular defect based on surface mesh deformation},
  author={Du, Wen and Wang, Hao and Zhao, Chenche and Cui, Zhiming and Li, Jiaqi and Zhang, Wenbo and Yu, Yao and Peng, Xin},
  journal={Journal of Stomatology, Oral and Maxillofacial Surgery},
  volume={125},
  number={5},
  pages={101973},
  year={2024},
  publisher={Elsevier}
}

@article{fang2024correspondence,
  title={Correspondence attention for facial appearance simulation},
  author={Fang, Xi and Kim, Daeseung and Xu, Xuanang and Kuang, Tianshu and Lampen, Nathan and Lee, Jungwook and Deng, Hannah H and Liebschner, Michael AK and Xia, James J and Gateno, Jaime and others},
  journal={Medical image analysis},
  volume={93},
  pages={103094},
  year={2024},
  publisher={Elsevier}
}

@inproceedings{yang2025blending,
  title={Blending Parametric Model and Neural Refinement: A Coarse-to-Fine Approach for Predicting Facial Changes in Orthognathic Surgery},
  author={Yang, Jiawen and He, Dongming and Tian, Xuanyu and Ma, Lei and Yu, Jingyi and Wang, Xudong and Zhang, Yuyao},
  booktitle={2025 IEEE 22nd International Symposium on Biomedical Imaging (ISBI)},
  pages={1--5},
  year={2025},
  organization={IEEE}
}

\end{document}